\documentclass[sigconf,10pt]{acmart}

\usepackage{pifont}
\usepackage{balance}
\usepackage{enumitem}
\usepackage{graphicx}
\usepackage{subfigure}
\usepackage{multirow}

\usepackage{tablefootnote}
\usepackage{threeparttable}

\usepackage[ruled, vlined, linesnumbered, ruled]{algorithm2e}
\usepackage{amsmath}

\usepackage{color}
\usepackage[noend]{algpseudocode}
\usepackage{xspace}
\usepackage{fancyhdr}
\usepackage{multicol}
\usepackage{amsfonts}
\usepackage{ulem}

\usepackage{makecell}
\usepackage{graphicx}
\usepackage{tikz}
\newcommand*{\circled}[1]{\lower.8ex\hbox{\tikz\draw (0pt, 0pt)%
    circle (.47em) node {\makebox[0.4em][c]{\small #1}};}}

\def\ie{\textit{i.e.}\xspace}

\def\method{\textsf{Pallas}\xspace}
\def\RC{\textsf{Recomputation}\xspace}
\def\FC{\textsf{Full-Copy}\xspace}
\def\Detour{\textsf{Detour}\xspace}
\def\ctHO{\textsf{ctHO}\xspace}

\AtBeginDocument{
  \providecommand\BibTeX{{%
    \normalfont B\kern-0.5em{\scshape i\kern-0.25em b}\kern-0.8em\TeX}}}

\setcopyright{none}

\acmDOI{}

\acmISBN{}

\begin{document}
 


\title{Pallas: A Proactive KV Cache Migration Framework for LLM Inference in AI-RAN}

\author{Tianhang Ding, Jianchun Liu, Hongli Xu \\
University of Science and Technology of China}


\begin{abstract}
AI-RAN brings large language model (LLM) serving close to mobile users, but cellular handover can separate an active request from its inference state: the user attaches to a target base station (gNB) while the large and growing key-value (KV) cache remains at the source.  
Retaining inference at the source preserves service continuity but persistently increases inter-token latency (ITL), whereas recovering the state at the target restores serving locality but requires KV-cache transfer, recomputation, or a combination of both only after handover, directly prolonging service interruption time (SIT).

This work presents \method, a \textit{proactive} KV-cache migration framework that prepares the inference state at the predicted target before handover, in parallel with ongoing source-side inference and token delivery.
At the preparation trigger, \method partitions the token sequence into a stable historical prefix and an evolving suffix.
The target reconstructs the prefix through local prefill, while the source streams the KV blocks generated for the suffix.
At handover, the target assembles both portions into an up-to-date KV cache and resumes decoding locally, leaving only unfinished preparation to contribute to SIT.
An online scheduler selects the \textit{prefetching window}, which determines how early preparation begins before handover, based on mobility predictions and runtime telemetry.
Across three LLMs and $100$--$500~\mathrm{Mbps}$ inter-gNB links, our vLLM-based prototype reduces average SIT by factors of $2.28$--$89.68$ over target-side recovery approaches and lowers average ITL by $16.0\%$--$50.0\%$ compared with source-side forwarding.


\end{abstract}



\maketitle

\pagenumbering{arabic}
\fancyfoot[C]{\fontsize{10pt}{40pt}\selectfont\thepage}
\setcounter{page}{1}
\thispagestyle{fancy}
\section{Introduction}\label{sec_introduction}
Radio Access Networks (RANs) are evolving from connectivity infrastructure into distributed computing platforms, with AI-RAN colocating accelerators and RAN functions at or near base stations (gNBs) \cite{10024837,kundu2025ai,kholmatov2025aora}. 
This proximity shortens the communication path to mobile users, making AI-RAN attractive for latency-sensitive large language model (LLM) services that continuously stream generated tokens \cite{298687, servegen}. 
However, \textit{user mobility} disrupts this serving model: after handover, the user equipment (UE) attaches to a target gNB while its service state remains at the source gNB \cite{abdah2020handover,ahmad2025edgewarp}. 
For LLM inference, this state includes an ever-growing KV-cache required to generate subsequent tokens \cite{kwon2023efficient}. 
Mobility can therefore separate an ongoing request from the state needed to continue it, undermining the latency advantage of AI-RAN.

Restoring LLM serving locality after this separation is challenging because the inference state is both large and continuously evolving.
During autoregressive decoding, the KV cache stores the key and value tensors of every processed token across all transformer layers and grows linearly with context length \cite{kwon2023efficient, 10.1145/3651890.3672274}.
For Qwen3-32B~\cite{yang2025qwen3}, an FP16 or BF16 KV cache
grows by approximately $256~\mathrm{KiB}$ per token\footnote{
The cache size is
$2\times64\times8\times128\times2=262{,}144$ bytes per token,
where the factors correspond to the key/value tensors, 64 layers,
8 KV heads, a head dimension of 128, and two bytes per element.
}, reaching
approximately $1~\mathrm{GiB}$ at $4{,}000$ tokens.
At a nominal inter-gNB bandwidth of $300~\mathrm{Mbps}$,
transmitting the payload alone takes approximately 28 seconds.
Thus, handover creates a pronounced timescale mismatch, \ie, the UE's network attachment can change rapidly, whereas relocating the inference state required to continue its inference session may take orders of magnitude longer.

This timescale mismatch creates a fundamental tension between \textit{service continuity} and \textit{serving locality}.
To avoid relocating the KV cache, \Detour \cite{ai_ran_wg3_2026} retains inference at the source gNB and forwards the generated tokens to the UE through the target gNB after handover. 
Although this approach keeps the accumulated inference state available at the source and therefore avoids LLM-state recovery, it introduces additional routing hops and persistently degrades inter-token latency (ITL).
As shown in Figure~\ref{fig:pre_exp1}, additional forwarding hops progressively shift the ITL distribution toward higher latency and substantially worsen its tail. 
Under the three-hop setting with Qwen3-32B, the P90 and P99 ITLs reach 126.68~ms and 263.54~ms, respectively, exceeding the 100~ms latency objective \cite{mlperf_inference_rules}.
\Detour therefore preserves service continuity only by sacrificing serving locality and sustained token-delivery latency.

In contrast, restoring serving locality requires making the inference state available at the target gNB. 
A straightforward approach, \textit{\FC} \cite{sun2024llumnix}, transfers the complete KV cache, following the basic mechanism of live KV-cache migration in data-center LLM serving systems.
However, over bandwidth-constrained inter-gNB links, the target cannot resume decoding until the entire, potentially multi-gigabyte KV cache has been transferred, placing the full transfer delay on the post-handover critical path and preventing the UE from receiving any newly generated tokens during this interval.
\textit{\RC} \cite{fu2024serverlessllm} avoids transferring the cache itself by sending the much smaller token sequence and rebuilding the cache through target-side prefill. 
Nevertheless, the target must replay the entire accumulated context before it can generate even the first post-handover token, causing the service interruption to grow with context length and consuming substantial target-side GPU computation precisely when the request needs to resume.

The most closely related approach, ctHO \cite{lee2026lowlatencyedgellmhandover}, combines target-side token prefill with direct KV-cache transfer. 
Specifically, the target reconstructs a prefix of the accumulated KV cache from historical tokens, while the source transfers the remaining KV-cache suffix. 
By jointly selecting this prefix--suffix partition and allocating backhaul bandwidth across multiple UEs undergoing handover, ctHO balances the completion times of target-side prefix reconstruction and source-to-target suffix transfer, thereby minimizing the worst-user handover delay.
However, ctHO assumes that token streaming is suspended at handover and places both target-side prefill and suffix transfer after that point.
Consequently, even with an optimized prefix--suffix partition, target-side decoding must wait for both recovery paths to finish, and the slower path directly determines the resulting service interruption time (SIT). 
In conclusion, \FC, \RC, and ctHO explore different combinations of communication and computation, but share the same temporal structure, \ie, \textit{state recovery begins only after handover}. 

This shared limitation motivates treating handover as a deadline for target-side state preparation rather than as the trigger for recovery.
Prior studies have shown that radio measurements, mobility trajectories, and association histories can predict the likely target gNB and approximate handover time \cite{abdah2020handover,mei2022realtime,ahmad2025edgewarp}.
Such advance estimates allow the target to prepare the inference state before handover while the UE remains connected to the source and continues receiving generated tokens.
However, exploiting this opportunity is not as simple as prefetching a static object: the KV cache grows with every token generated at the source, so the target must prepare the historical state while continuously incorporating newly generated state.

To this end, we propose \method, a proactive KV-cache migration framework for low-latency LLM serving in AI-RAN.
At the preparation trigger, \method marks the block-aligned split position as a split point: all tokens processed by that time form a stable prefix, while tokens generated afterward form an evolving suffix.
The target gNB receives the compact token sequence of the stable prefix and reconstructs its KV cache through local prefill.
In parallel, the source gNB continues generating and delivering tokens to the UE while streaming the KV blocks of the evolving suffix to the target. 
The target then combines the reconstructed prefix with the received suffix to obtain a complete, up-to-date KV cache. 
After handover, local decoding can resume as soon as both portions are available, so only unfinished preparation contributes to SIT. 
In essence, \method reconstructs what is already stable and transfers only what continues to change.

We define the interval between the preparation trigger and the predicted handover as the \textit{prefetching window}. 
A short window leaves prefix reconstruction or suffix synchronization unfinished, causing the remaining work to contribute directly to SIT.
Conversely, an excessively long window establishes an earlier split point, producing a larger incremental suffix, significantly increasing KV-cache transfer volume.
Importantly, finding a proper window is nontrivial because the trigger time jointly determines the sizes of the historical prefix and evolving suffix, whose completion times depend on the context length, target-side prefill throughput, source-side decode throughput, and inter-gNB goodput.
Because these factors differ across requests and change over time, no fixed window can consistently balance service interruption against preparation overhead.
Mobility-prediction updates further complicate this decision, \ie, an updated handover time changes the remaining preparation time, while a new target prediction can invalidate the state already prepared.

The main contributions of this paper are as follows.
\begin{itemize}[topsep=2pt]
    \item We reveal the continuity--locality trade-off in mobile LLM handover, showing that existing strategies incur either persistent detour latency or post-handover state-recovery delay, which motivates treating handover as a preparation deadline rather than a recovery trigger.
    \item We design \method, a proactive KV-cache migration framework that reconstructs the stable historical prefix at the target in parallel with source-side decoding and incremental suffix streaming, thereby moving most state-recovery work before handover.
    \item We formulate prefetching-window selection based on prefill/decoding rates and inter-gNB goodput, and develop an online scheduler that adapts the trigger to changing system conditions and mobility predictions while bounding stale-preparation duration.
    \item We implement \method on vLLM 0.8.5 and deploy it across two physical servers, each equipped with a single NVIDIA RTX A6000 GPU and connected by 1-Gbps Ethernet. 
    Prototype experiments and trace-driven repeated-handover simulations across three LLMs and diverse inter-gNB conditions show that \method substantially reduces SIT relative to baselines while avoiding the persistent ITL overhead.
\end{itemize}

\section{Background and Motivation}\label{sec:prelim}

\subsection{LLM Serving under AI-RAN Handover}
\label{sec:preliminaries}
We consider an AI-RAN system in which GPU-equipped edge servers are colocated with gNBs to serve mobile UEs. 
During an active LLM session, the UE is connected to a serving gNB over the radio access link, while the colocated inference engine executes the request and returns generated tokens directly to the UE. 
This proximity shortens the service path and reduces communication latency \cite{3gpp_edge_computing,kundu2025ai,kholmatov2025aora}. 
When the UE moves across a cell boundary, the RAN performs a handover that redirects its network attachment from the current serving gNB (or source) to a new serving gNB (or target) \cite{abdah2020handover}. 
Handover updates the UE's access path and maintains network connectivity, but does not automatically relocate the application-layer state of an ongoing request \cite{ahmad2025edgewarp}. 
We assume that the source and target can execute the same LLM configuration, and our focus is the per-request inference state that must be preserved across handover.

An LLM request proceeds through two inference phases: prefilling and decoding \cite{298687}. 
During prefilling, the inference engine processes the input prompt and constructs the corresponding key-value (KV) cache. 
It then enters autoregressive decoding, where each new token is generated using the cached key and value tensors of all preceding tokens. 
The KV cache avoids repeatedly processing the complete token history, but grows whenever a new token is generated \cite{kwon2023efficient, 10.1145/3651890.3672274}. 
Consequently, it is both indispensable for continuing the request and continuously evolving throughout the decoding process.
For an LLM with $L$ transformer layers, $H_{\mathrm{KV}}$ KV heads, head dimension $d_h$, and $b$ bytes per cache element, the per-token KV-cache footprint is
\begin{equation}
    s_{\mathrm{KV}}=2\cdot L\cdot H_{\mathrm{KV}}\cdot d_h\cdot b,\qquad 
    S_{\mathrm{KV}}(N)=N\cdot s_{\mathrm{KV}},
\label{eq:s_kv}
\end{equation}
where the factor of two accounts for the key and value tensors, and $S_{\mathrm{KV}}(N)$ denotes the cache size after $N$ tokens. 

During decoding, this request-specific state resides in the GPU memory of the source gNB. After handover, however, the UE communicates through the target gNB while its accumulated KV cache remains at the source. The target therefore lacks the state required to immediately continue decoding, even though the UE's network attachment has already been updated. Continuing the session requires either retaining inference at the source and forwarding subsequent tokens through the target, or making the inference state available at the target through KV-cache transfer, recomputation, or a combination of both. The next subsection characterizes the limitations of these alternatives.


\begin{figure}[t]
\centering
\includegraphics[width=1\linewidth]{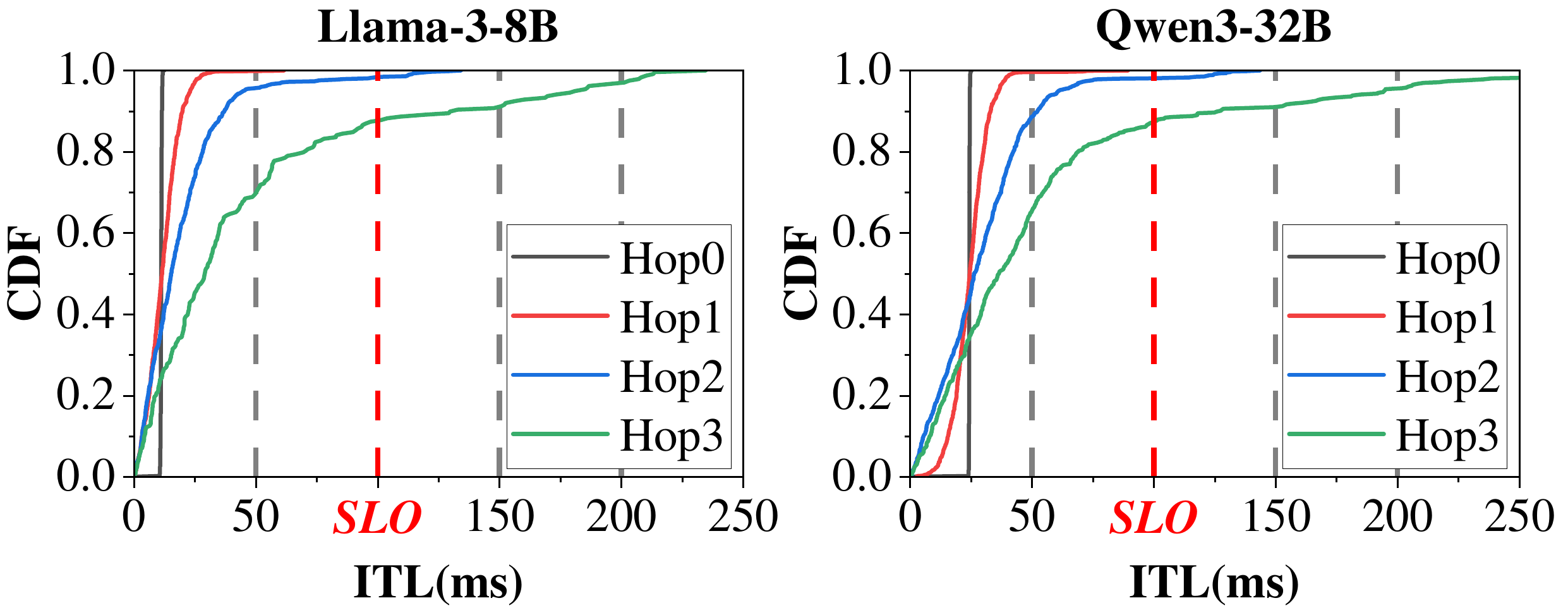}
\caption{CDF of ITL under the \Detour strategy. Increased routing hops lead to severe ITL degradation. The vertical red dashed line at 100 ms marks the SLO.}
\label{fig:pre_exp1}
\vspace{-5mm}
\end{figure}

\subsection{Limitations of Existing Strategies} 
\label{sec:bottlenecks}
To quantify the limitations of existing methods, we conduct a set of preliminary experiments that examine the latency overhead of continuing inference at the source gNB and the service interruption caused by KV-cache recovery at the target gNB.

\textbf{Source-side retention.}
\Detour \cite{ai_ran_wg3_2026} retains inference at the source gNB after handover and uses the target gNB as a relay for every subsequently generated token. 
Although this strategy avoids recovering the KV cache at the target, it lengthens the serving path throughout the remainder of the request. 
To quantify this persistent cost, we measure ITL while varying the number of forwarding hops between the inference engine and the UE.
As shown in Figure~\ref{fig:pre_exp1}, the ITL distributions of both Llama-3-8B and Qwen3-32B progressively shift toward higher latency and develop substantially longer tails as the routing path grows.
Under the three-hop setting, their P90 ITLs reach $130.54~\mathrm{ms}$ and $126.68~\mathrm{ms}$, respectively; the P99 ITL of Qwen3-32B further reaches $263.54~\mathrm{ms}$. 
Following the conversational TPOT bound adopted by MLPerf \cite{mlperf_inference_rules}, we use $100~\mathrm{ms}$ as the token-latency service level objective (SLO) throughout the evaluation. 
The fraction of individual ITL samples exceeding this threshold increases from $1.74\%$ and $1.93\%$ under two hops to $12.61\%$ and $12.94\%$ under three hops for Llama-3-8B and Qwen3-32B, respectively. 
Unlike a one-time migration cost, this path-induced penalty affects every token generated after handover.

\noindent\textbf{Observation 1.} 
\Detour avoids target-side state recovery by converting it into a persistent token-delivery penalty that worsens with routing distance.

\textbf{Target-side recovery.}
Restoring serving locality requires making the accumulated KV cache available at the target before local decoding can resume.
We next characterize the communication and computation costs of representative target-side recovery strategies.
\FC \cite{sun2024llumnix} transfers the complete KV cache from the source to the target.
Its recovery time therefore grows with cache size and is inversely proportional to the inter-gNB goodput available to the migration.
As shown in Table~\ref{tab:pre_exp3}, at 300~Mbps inter-gNB goodput, the SIT of \FC increases from 7.17~s at a 1K-token context to 29.07~s at 4K tokens.
Because transfer starts only after handover, the target cannot resume decoding throughout this interval.

\RC \cite{fu2024serverlessllm} instead sends the compact token sequence and reconstructs the KV cache through target-side prefill.
This avoids transferring the cache itself but places the entire prefill computation on the interruption path.
Table~\ref{tab:pre_exp3} shows that the SIT of \RC increases from 581~ms at a 1K-token context to 2,103~ms at 4K tokens, corresponding to an approximately 3.6$\times$ increase as the context grows.
Thus, although \RC replaces KV-cache transfer with local computation, its increasing recomputation delay remains entirely on the post-handover interruption path before the UE can receive the first token from the target.

The most closely related strategy, \ctHO \cite{lee2026lowlatencyedgellmhandover}, combines the two recovery operations: the target reconstructs a KV-cache prefix through token prefill, while the source transfers the remaining suffix.
By optimizing the prefix--suffix partition, \ctHO balances the completion times of recomputation and transfer and reduces handover delay relative to using either operation alone.
Nevertheless, Table~\ref{tab:pre_exp3} shows that the SIT of \ctHO still reaches 2,058~ms at a 4K-token context, despite its optimized prefix--suffix partition.
Both operations are placed after handover, and decoding cannot resume until they have both completed.
Consequently, the entire recovery interval remains exposed as SIT.

\noindent\textbf{Observation 2.} \FC, \RC, and \ctHO explore different combinations of communication and computation, but share the same reactive timing, \ie, their state-recovery work begins only after handover and therefore directly contributes to SIT.


\begin{table}[t]
\centering
\caption{Target-side recovery time (ms) of representative recovery strategies with Qwen3-32B at 300~Mbps inter-gNB goodput.}
\label{tab:pre_exp3}
\small
\setlength{\tabcolsep}{5pt}
\begin{tabular}{@{}lccc@{}}
\toprule
\textbf{Strategy} & \textbf{1K} & \textbf{2K} & \textbf{4K} \\
\midrule
\FC   & 7165 & 14310 & 29066 \\
\RC   & 581  & 1208  & 2103  \\
\ctHO & 554  & 1119  & 2058  \\
\bottomrule
\end{tabular}
\vspace{-5mm}
\end{table}

\subsection{Opportunities for System Design}
\label{sec:Optimal_Timing}

The reactive timing identified in \S~\ref{sec:bottlenecks} suggests a temporal opportunity: instead of treating handover as the point at which recovery begins, the system can use it as a deadline by which state preparation should be largely completed.
Prior studies have shown that radio measurements, mobility trajectories, and association histories can be used to predict both the next serving cell and the handover time \cite{abdah2020handover,mei2022realtime,ahmad2025edgewarp}.
We denote the predicted target gNB and the predicted handover time by $\hat{g}_{\mathrm{tar}}$ and $\hat{t}_{\mathrm{HO}}$, respectively. 
If preparation is triggered at time $t_{\mathrm{trig}}$, the resulting \textit{prefetching window} is
\begin{equation}
T_w = \hat{t}_{\mathrm{HO}} - t_{\mathrm{trig}}.
\label{eq:window}
\end{equation}
During this interval, the UE remains connected to the source, which continues generating and delivering tokens, while the predicted target prepares the inference state in parallel. 

\textbf{Preparing an evolving inference state.} 
Proactive preparation is more difficult than prefetching a static object because the KV cache continues to grow while the source serves the UE. 
At the trigger time, the accumulated token sequence forms a stable prefix whose KV cache can be reconstructed at the target through prefill. 
Tokens generated afterward form an evolving suffix whose KV blocks must be continuously incorporated into the target state. 
Let $r_{\mathrm{d}}$ denote the per-request decode throughput at the source gNB.
The incremental KV-cache volume produced over a window of length $T_w$ is approximately
\begin{equation}
    S_{\mathrm{inc}}(T_w)\approx r_{\mathrm{d}}\cdot T_w\cdot s_{\mathrm{KV}},
\label{eq:S_inc}
\end{equation}
A one-time snapshot becomes stale as decoding proceeds, whereas pausing the source to preserve a static state would itself interrupt token delivery. 
A viable \textit{proactive} design thus reconstructs the stable history while tracking newly generated state without suspending source-side inference.

\begin{figure}[t]
\centering\includegraphics[width=\linewidth]{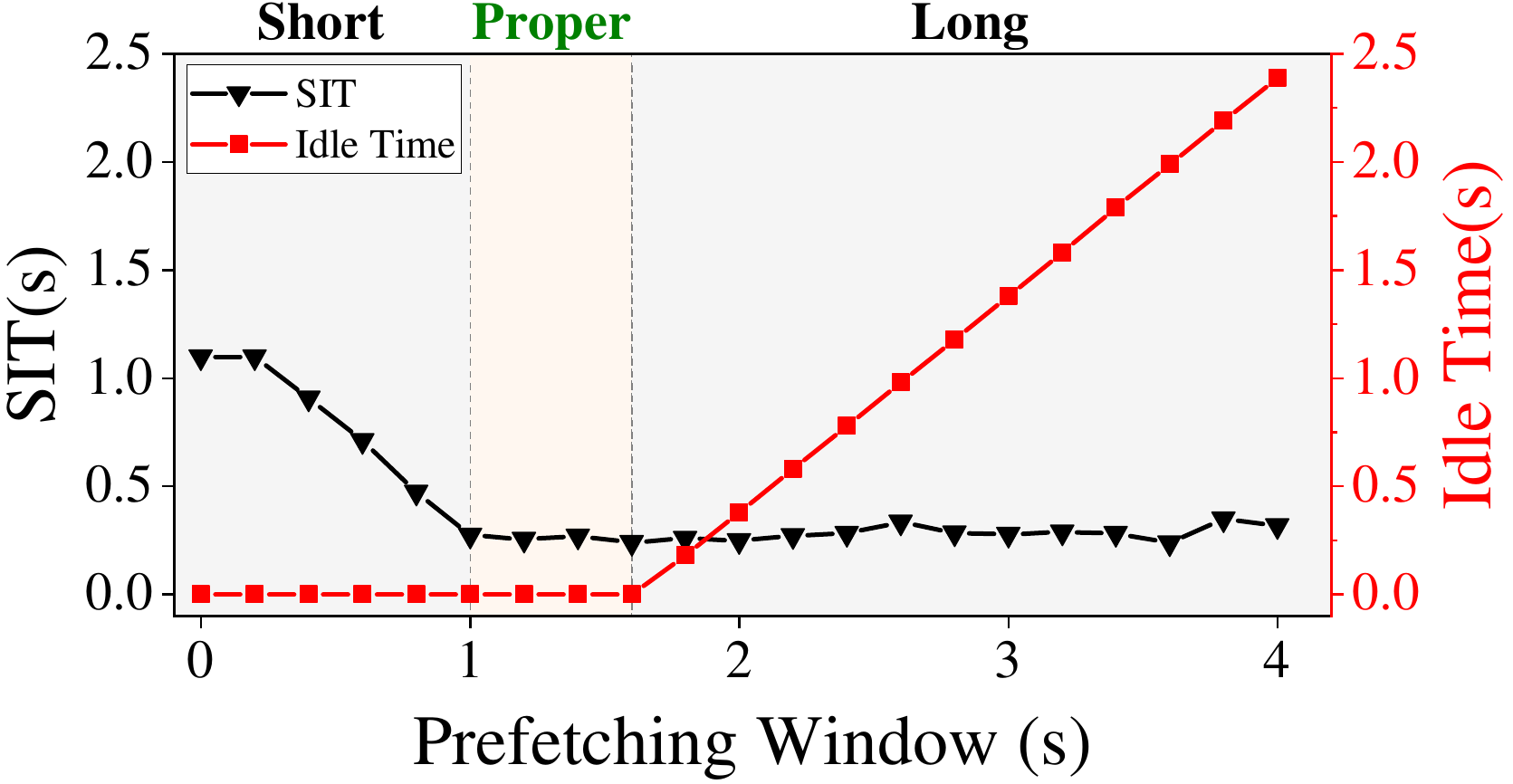}
\caption{Prefetching-window trade-off under a representative operating condition (300~Mbps inter-gNB goodput, Qwen3-14B, and a 3,000-token context).}
\label{fig:window-size}
\vspace{-5mm}
\end{figure}

\textbf{Selecting the prefetching window.}
The trigger time determines both how much preparation is completed before handover and how early the target-side state is prepared.
To illustrate this trade-off, we vary \(T_w\) under a representative setting of 300~Mbps inter-gNB goodput, Qwen3-14B, and a 3,000-token context.
As shown in Figure~\ref{fig:window-size}, increasing \(T_w\) from 0 to about 1~s reduces SIT by approximately 77\%, from 1.1~s to 0.25~s, while the target-side idle time remains negligible.
Within an intermediate window range, both SIT and idle time remain low.
Beyond this range, further increasing \(T_w\) provides little additional SIT reduction but causes the idle time to grow rapidly, reaching approximately 2.4~s at \(T_w=4~\mathrm{s}\), equivalent to 60\% of the entire prefetching window.
This illustrates the two-sided trade-off: a short window leaves residual preparation on the handover critical path, whereas a long window triggers preparation unnecessarily early.
Hence, effective proactive migration requires selecting a \textit{proper prefetching window} that balances residual service interruption against early target-side idle time.

\section{System Overview of Pallas}\label{sec:overview}
\begin{figure}[t]
\centering
    \includegraphics[width=1.0\linewidth]{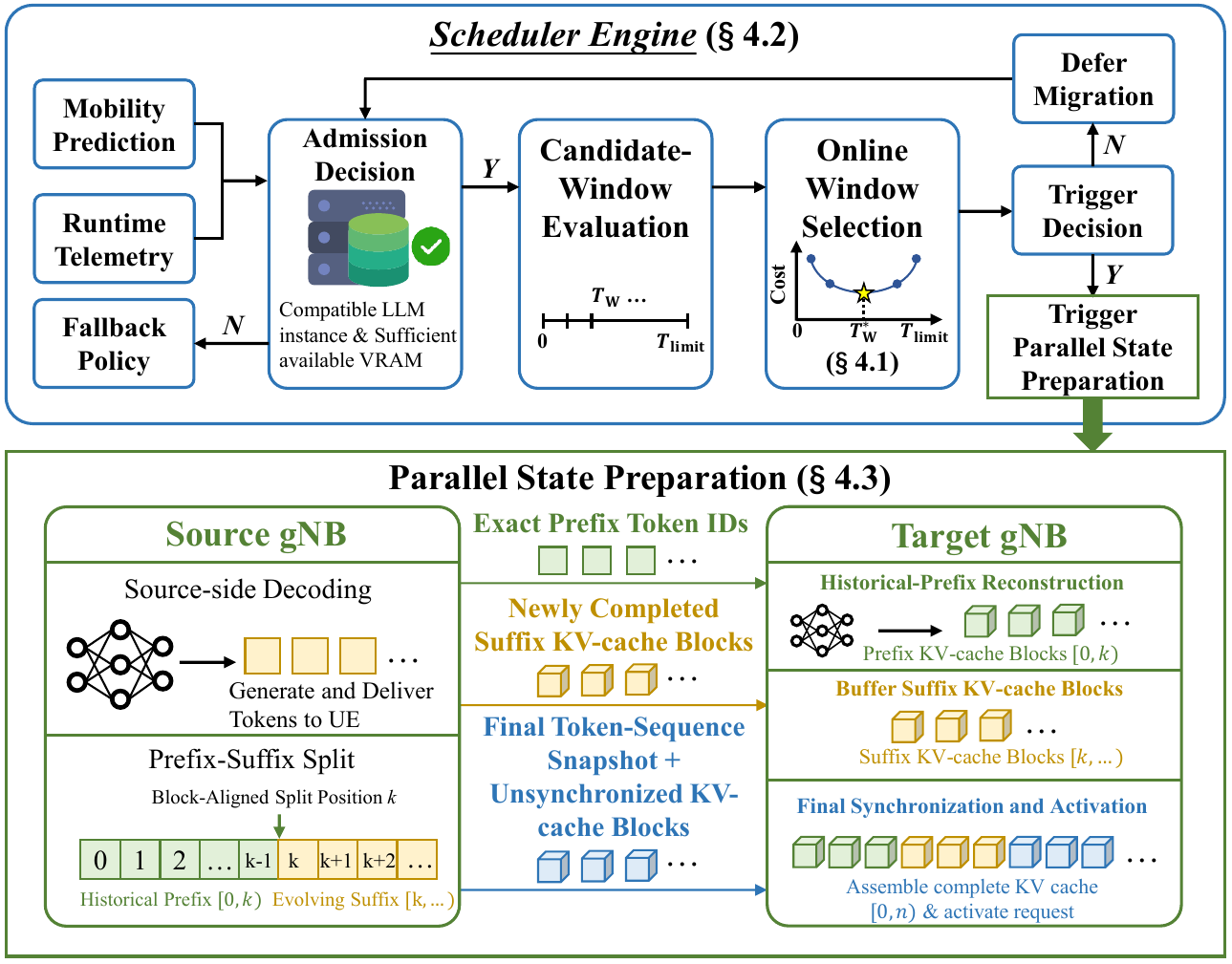}
    \caption{The workflow of \method.}
    \label{fig:overview}
    \vspace{-5mm}
 \end{figure}

To exploit the aforementioned opportunities, we propose \method, which proactively moves inference-state preparation into the prefetching window and executes it in parallel with ongoing source-side inference.
Figure~\ref{fig:overview} provides an overview of its online decision and parallel state-preparation workflow.
At each control cycle, the \textit{Scheduler Engine} of \method combines mobility predictions with the current context length and runtime estimates of target-side prefill throughput, source-side per-request decode throughput, and inter-gNB goodput.
It first verifies that the predicted target can execute a compatible LLM instance and has sufficient available VRAM.
If the predicted target is infeasible, \method invokes a fallback policy that temporarily retains source-side service or redirects preparation to a feasible serving tier.
If the target is feasible, the Scheduler Engine evaluates candidate prefetching windows and selects the window with the lowest estimated cost.
If the predicted remaining time to handover is still longer than the selected window, migration is deferred until the next control cycle.
Otherwise, \method triggers parallel state preparation at the predicted target.
Once triggered, the source records the block-aligned split position as the prefix--suffix split and sends the exact historical-prefix token IDs to the target, which reconstructs the corresponding KV cache through local prefill.
Meanwhile, the source remains the only active decoder, continues generating and delivering tokens to the UE, and asynchronously streams newly completed suffix KV-cache blocks to the target.
Thus, historical-prefix reconstruction, source-side decoding, and suffix transfer proceed in parallel during the prefetching window.
At handover, the source records the final sequence position and sends the final token-sequence snapshot together with any unsynchronized KV-cache blocks.
Once prefix reconstruction is complete and the state for all sequence positions is available, the target combines the reconstructed prefix with the received suffix and activates the request for local decoding.
Before preparation begins, updated mobility predictions and runtime telemetry are incorporated into subsequent window decisions.
After preparation begins, a revised handover time changes the preparation deadline without invalidating the existing state, whereas a change in the predicted target cancels the obsolete preparation and triggers a new admission and window decision.

\begin{figure}
    \centering
    \includegraphics[width=\linewidth]{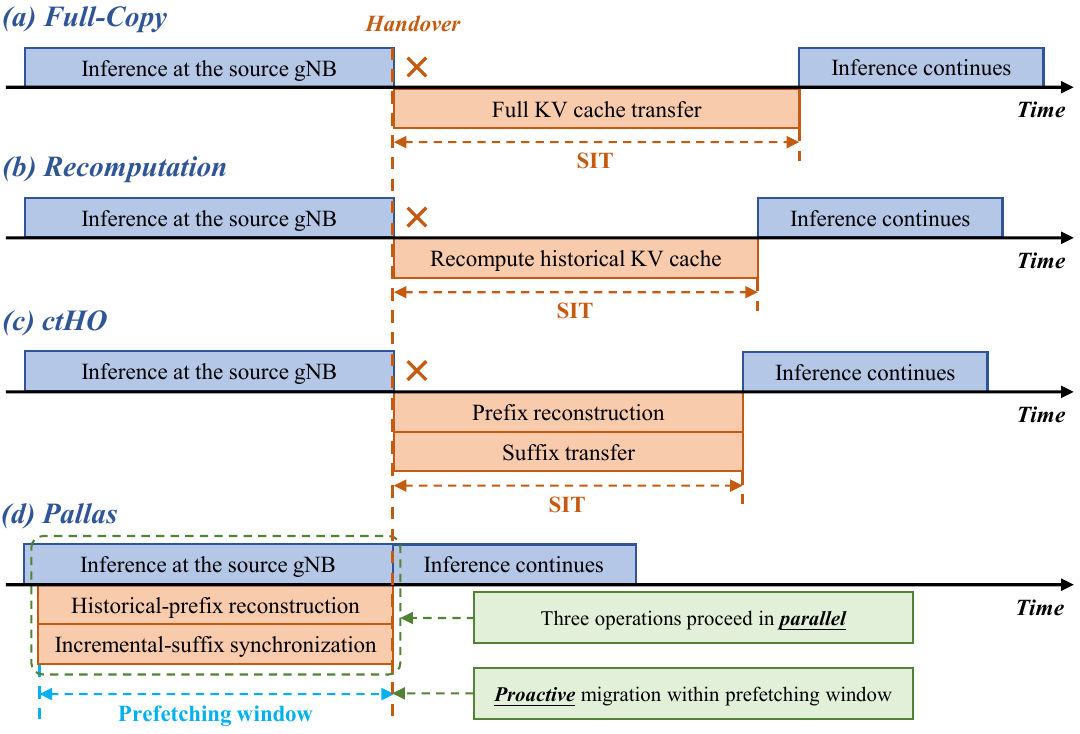}
    \caption{KV-cache recovery timelines of \FC, \RC, \ctHO, and \method during handover.}
    \label{fig:Comparison}
\vspace{-5mm}
\end{figure}

Figure~\ref{fig:Comparison} further illustrates how \method differs from existing KV-cache recovery strategies in terms of execution timing.
As shown in Figure~\ref{fig:Comparison}(a), \FC postpones the entire KV-cache transfer until after handover, placing the full transfer delay on the post-handover critical path.
\RC replaces KV-cache transfer with target-side historical KV-cache reconstruction, but Figure~\ref{fig:Comparison}(b) shows that the complete recomputation still remains on the post-handover path.
\ctHO further overlaps prefix reconstruction with suffix transfer, as illustrated in Figure~\ref{fig:Comparison}(c), thereby reducing the recovery delay to the completion time of the slower operation.
However, both operations begin only after handover, leaving target-side decoding blocked by state recovery.
In contrast, \method moves state preparation ahead of handover, as shown in Figure~\ref{fig:Comparison}(d).
Historical KV-cache reconstruction at the target and incremental KV-cache transfer from the source proceed within the prefetching window in parallel with ongoing source-side inference.
As a result, most state-preparation work can be completed before handover.

\section{System Design of Pallas}\label{sec:design}
\subsection{Prefetching-Window Trade-off Model}
\label{Sec:modeling}

As discussed in \S~\ref{sec:Optimal_Timing}, the prefetching-window length \(T_w\) controls two opposing effects.
A short window may leave state-recovery work unfinished at handover, whereas an excessively long window initiates preparation earlier than necessary and keeps the reconstructed state at the target for longer.
We make this trade-off explicit by modeling the expected computation and communication progress within a candidate window.

\noindent\textbf{Historical-prefix reconstruction.}
At the preparation trigger, \method splits the token sequence at the block-aligned split position.
All tokens processed by that time form a stable historical prefix, while subsequently generated tokens form an evolving suffix.
Let \(L_{\mathrm{hist}}\) denote the number of tokens in this prefix, and let $r_p$ denote the expected prefill throughput available to the migration session at the target gNB.
Unlike the standalone peak throughput, $r_p$ accounts for the computation capacity expected to be available under the current target-side load.
For an isolated migration, it corresponds to the target's measured prefill throughput, whereas under resource contention it reflects the effective throughput expected to be available to the migration session.
Let \(T_{\mathrm{load}}\) denote the latency of loading the model weights from host memory to the target GPU \cite{318391}.
The target-side time required to reconstruct the historical KV cache is:
\begin{equation}
      T_{\mathrm{hist}}=
    T_{\mathrm{load}}+\frac{L_{\mathrm{hist}}}{r_{\mathrm{p}}}.
    \label{eq:ready}
\end{equation}
If the required model is already resident on the target GPU, as in our prototype evaluation, then \(T_{\mathrm{load}}=0\).

\noindent\textbf{Incremental-suffix synchronization.}
While the target reconstructs the historical prefix, the source continues decoding and streams newly completed KV-cache blocks to the target.
As defined in Eq.~\eqref{eq:S_inc}, the resulting incremental KV-cache volume over a window of length \(T_w\) is \(S_{\mathrm{inc}}(T_w)\).
Let \(B\) denote the expected inter-gNB goodput available to the migration session.
Because the incremental KV cache is streamed while it is generated, only the portion that cannot be delivered within the prefetching window remains on the handover critical path.
The corresponding residual transmission delay is
\begin{equation}
    T_{\mathrm{res}}(T_w) = \max\left\{0, \frac{S_{\mathrm{inc}}(T_w)}{B} - T_w\right\}.
    \label{eq:residual}
\end{equation}
Here, \(B\) is the effective goodput observed by the migration session and therefore accounts for its available link capacity and protocol overhead.

\noindent\textbf{Service interruption.}
Target-side decoding can resume only after both the historical KV cache and the incremental suffix have become available.
Because historical-prefix reconstruction and suffix transmission proceed in parallel, the slower unfinished operation determines the migration-induced delay after handover.
Let \(T_0\) denote the window-independent activation latency after the required state becomes available, including final state assembly, request activation, and the first target-side decoding step.
The resulting service interruption is
\begin{equation}
    T_{\mathrm{SIT}}(T_w) = T_0 + \max
    \left\{
        0,\,
        T_{\mathrm{hist}}-T_w,\,
        T_{\mathrm{res}}(T_w)
    \right\}.
    \label{eq:SIT}
\end{equation}
Thus, if both preparation paths complete before handover, the resulting service interruption is reduced to \(T_0\).

\noindent\textbf{Early-preparation exposure.}
A longer window may allow historical recomputation to finish well before handover.
During this interval, the reconstructed historical KV cache remains resident at the target even though the UE is still served by the source.
We quantify this early-preparation exposure as
\begin{equation}
    T_{\mathrm{early}}(T_w)
    =
    \max
    \left\{
        0,\,
        T_w-T_{\mathrm{hist}}
    \right\}.
    \label{eq:early}
\end{equation}
\(T_{\mathrm{early}}\) is not GPU idle time: the target GPU may continue serving other requests after completing the prefill.
Instead, it measures how long migration-specific state is committed before it is needed and serves as a proxy for premature resource commitment and exposure to wasted preparation if the predicted target changes.

\method balances post-handover interruption against early preparation using the following objective:
\begin{equation}
    \mathcal{J}(T_w)
    =
    \alpha\cdot T_{\mathrm{SIT}}(T_w)
    +
    (1-\alpha)\cdot T_{\mathrm{early}}(T_w),
    \qquad 0\le\alpha\le1,
    \label{eq:cost}
\end{equation}
where a larger \(\alpha\) prioritizes reducing service interruption, whereas a smaller \(\alpha\) discourages unnecessarily early preparation.
The objective captures the latency consequences of incremental KV-cache transfer rather than directly penalizing its total volume.
Although a longer window produces more incremental KV-cache blocks, these blocks are transmitted concurrently with source-side decoding.
Only the backlog remaining at handover contributes to \(T_{\mathrm{SIT}}\) through \(T_{\mathrm{res}}(T_w)\).
When evaluating a candidate window before handover, \(L_{\mathrm{hist}}\) is not yet observed because the source continues decoding until the candidate trigger time.
The online procedure in \S~\ref{sec:window_decision} therefore predicts the candidate-specific historical-prefix length and evaluates Eq.~\eqref{eq:cost} using runtime estimates of $r_p$, $r_d$, and \(B\).

\subsection{Online Prefetching-Window Selection}
\label{sec:window_decision}

The model in \S\ref{Sec:modeling} quantifies the cost of a given prefetching window, but selecting the window online introduces two practical challenges. 
First, the historical-prefix length at a candidate trigger time is not yet known because decoding continues until preparation begins. 
Second, the resources available to migration vary with target-side load and network conditions. 
\method addresses these challenges with a lightweight Scheduler Engine that periodically combines mobility predictions with runtime telemetry and determines whether preparation should be triggered.

\textbf{Mobility and telemetry inputs.}
\method treats mobility prediction as an external input rather than requiring a particular prediction algorithm. 
At each control cycle, the predictor provides the predicted target gNB $\hat{g}_{\mathrm{tar}}$ and handover time $\hat{t}_{\mathrm{HO}}$. 
Given the current time $t_{\mathrm{curr}}$, the predicted remaining time is $\widehat{T}_{\mathrm{remain}}=\hat{t}_{\mathrm{HO}}-t_{\mathrm{curr}}$. 
The Scheduler Engine additionally collects the current context length $K_{\mathrm{curr}}$, target-side prefill throughput $r_\mathrm{p}$, source-side per-request decode throughput $r_\mathrm{d}$, and inter-gNB goodput $B$.
The Scheduler Engine uses session-level effective rates rather than peak hardware or link capacities. 
For example, $r_p$ reflects the prefill throughput available to the migrating request under the current target-side load, while $B$ reflects the goodput available after contention and protocol overhead. 
To avoid reacting to short-term fluctuations, \method smooths the instantaneous measurements using an exponentially weighted moving average. 
The resulting estimates allow the window decision to account for both handover--handover and handover--inference contention without assuming exclusive GPU or network resources.

\textbf{Admission decision.}
Before evaluating candidate windows, the Scheduler Engine verifies that the predicted target can execute a compatible LLM instance and has sufficient available VRAM for the request state expected at handover.
If either condition is not satisfied, \method does not initiate direct preparation at that gNB and invokes the fallback policy described in \S\ref{sec:overview}.
Model compatibility and VRAM availability are therefore treated as admission conditions, and window optimization is performed only for a feasible target.

\textbf{Candidate-window evaluation.}
\method limits the candidate window to $T_{\mathrm{limit}}=\min(T_{\max},\widehat{T}_{\mathrm{remain}})$, where $T_{\max}$ prevents preparation from starting excessively early. 
The Scheduler Engine enumerates candidate windows over $[0,T_{\mathrm{limit}}]$ with granularity $\Delta T$.
For a candidate $T_w$, preparation would begin after $\widehat{T}_{\mathrm{remain}}-T_w$ seconds. 
Because the source continues decoding during this interval, \method first predicts the historical-prefix length at the corresponding trigger time and then evaluates the cost model from \S\ref{Sec:modeling}:
\begin{equation}
\begin{aligned}
\widehat{L}_{\mathrm{hist}}(T_w)
&=K_{\mathrm{curr}}+r_d\cdot \bigl(\widehat{T}_{\mathrm{remain}}-T_w\bigr),\\ 
T_w^*&=\arg\min_{T_w \in \mathcal{W}}\mathcal{J}\bigl(T_w;\widehat{L}_{\mathrm{hist}}(T_w)\bigr),
\end{aligned}
\label{eq}
\end{equation}
where $\mathcal{W}$ is the set of candidate windows and $\mathcal{J}(\cdot)$ is the objective in Eq. \eqref{eq:cost}.
This evaluation explicitly captures that a later trigger produces a longer historical prefix, whereas an earlier trigger produces a longer incremental suffix.

\textbf{Trigger decision.}
If $\widehat{T}_{\mathrm{remain}}>T_w^*$, the Scheduler Engine defers migration and reevaluates the decision in the next control cycle.
Once $\widehat{T}_{\mathrm{remain}}\leq T_w^*$, it triggers parallel state preparation at the predicted target.
The source records its block-aligned split position as the prefix--suffix split and starts parallel state preparation.
Triggering is a one-time transition for the current target prediction; subsequent changes in the predicted handover time or target are handled through replanning or cancellation rather than by reversing the completed trigger.

Algorithm~\ref{alg} summarizes the online procedure. With a bounded grid search, each control cycle evaluates $\mathcal{O}(T_{\mathrm{limit}}/\Delta T)$ candidates. 
The cost evaluation involves only scalar arithmetic and completes within microseconds in our implementation, making its overhead negligible relative to the control period and LLM inference timescale.
\method uses the observed resource availability to decide when migration should begin, but does not jointly allocate GPU cycles or inter-gNB bandwidth among concurrent migrations. 
Such contention is reflected through the effective rates supplied to the Scheduler Engine, and general multi-user resource allocation remains outside the scope of this work.

\begin{algorithm}[t]
\caption{Online prefetching-window selection}
\label{alg}

\KwIn{ Mobility prediction
$(\hat{g}_{\mathrm{tar}},\hat{t}_{\mathrm{HO}})$,
current request state, and runtime telemetry}
\KwOut{
Selected window $T_w^*$ if the predicted target
is feasible
}
Update $\widehat{T}_{\mathrm{remain}}$ and smooth $r_p$, $r_d$, and $B$ \\
\If{the predicted target lacks the required model or VRAM}
{Invoke the fallback policy and \Return}

Set $T_{\mathrm{limit}} \gets\min(T_{\max},\widehat{T}_{\mathrm{remain}})$ \\
Construct candidate set $\mathcal{W}\gets{\{0,\Delta T,\ldots,T_{\mathrm{limit}}\}}$ \\
\ForEach{$T_w\in\mathcal{W}$}{
Predict $\widehat{L}_{\mathrm{hist}}(T_w)$\\
Evaluate $\mathcal{J}(T_w;\widehat{L}_{\mathrm{hist}}(T_w))$}
$T_w^*
\gets
\arg\min_{T_w\in\mathcal{W}}
\mathcal{J}\!\left(
    T_w;
    \widehat{L}_{\mathrm{hist}}(T_w)
\right)$\\
\eIf{$\widehat{T}_{\mathrm{remain}}\leq T_w^*$}{
    Trigger parallel state preparation at
    $\hat{g}_{\mathrm{tar}}$\; 
}{
    Defer migration until the next control cycle
}

\end{algorithm}

\subsection{Parallel State Preparation}
\label{sec:migration_protocol}

Once preparation is triggered, \method reconstructs the target-side inference state without suspending source-side token generation. 
The protocol consists of initialization, parallel preparation, and final synchronization.

\noindent\textbf{Initialization and parallel preparation.}
At the trigger time, the source remains the sole owner of the request and records its block-aligned split position $k$. 
Tokens in $[0,k)$ form an immutable historical prefix, whereas subsequently generated tokens form an evolving suffix. 
The source sends the exact prefix token IDs, split position, request identifier, and migration epoch to the target, which reconstructs the prefix KV cache through local prefill.
Meanwhile, the source continues generating and delivering tokens to the UE. 
Newly completed suffix KV-cache blocks are streamed asynchronously with their logical token positions and buffered at the target. 
Only completed blocks are transferred during decoding; any completed but unsent blocks remain in a sender-side queue, whose backlog at handover corresponds to the residual transfer delay in Eq.~\eqref{eq:residual}. 
Thus, prefix reconstruction, source-side decoding, and suffix transfer proceed in parallel.

\noindent\textbf{Final synchronization and activation.}
At handover, the source finishes its current decoding step and records the final sequence position $n$. 
It then sends the final token-sequence snapshot and any unsynchronized KV-cache blocks, including the residual boundary block. 
The target waits until prefix reconstruction is complete and state for all positions in $[k,n)$ is available. 
It then combines the reconstructed prefix and received suffix and activates the request in the decoding stage, generating token $n$ without repeating prefill over the complete context. 
Only preparation unfinished at handover contributes to SIT.

\noindent\textbf{State consistency.}
Before activation, \method verifies three conditions: the prefix is reconstructed from the exact token IDs supplied by the source; the assembled KV cache covers all logical positions in $[0,n)$ without gaps or conflicts; and the source remains the only active decoder until ownership is transferred to the target. 
The migration epoch prevents blocks from a cancelled preparation attempt from entering the active request state.

\subsection{Prediction-Aware Migration Replanning}
\label{sec:runtime_replanning}

Mobility predictions may change as the UE approaches a cell boundary. 
Before preparation begins, \method simply updates $(\hat{g}_{\mathrm{tar}},\hat{t}_{\mathrm{HO}})$ and reruns Algorithm~\ref{alg}. 
Because no state has been prepared, these updates incur only Schedule Engine overhead.
After preparation begins, a revised handover time does not invalidate existing state. 
If handover is predicted earlier, \method continues preparation against the shortened deadline, and unfinished work contributes to SIT. 
If handover is delayed, the source continues serving the UE and streaming incremental KV blocks; the reconstructed prefix remains valid, although it may remain resident at the target for longer.

A change in the predicted target invalidates preparation at the previous gNB. 
\method increments the migration epoch, cancels the obsolete attempt, and releases its reconstructed and buffered state. 
It then performs admission control and reruns the window decision for the new target. 
Because \method prepares at only one predicted target at a time, obsolete messages can be discarded using their migration epoch. 
If the actual handover target differs from the prepared target, \method falls back to post-handover recovery at the actual target rather than activating an inconsistent state. 
If the new target is infeasible, Pallas invokes the fallback policy described in \S\ref{sec:overview}.

We bound the duration of preparation discarded because of incorrect target predictions. 
Let $T_{\mathrm{avail}}$ denote the interval between the time when the target prediction becomes stably correct and the actual handover, with $T_{\mathrm{avail}}=0$ if this never occurs. 
Let $E_{\mathrm{early}}$ denote the maximum amount by which the predicted handover time precedes the actual handover before target stabilization. 
Because \method caps the prefetching window at $T_{\max}$, the discarded preparation duration satisfies 
\begin{equation}
T_{\mathrm{waste}}
\leq
\max\!\left\{T_{\max}+E_{\mathrm{early}}-T_{\mathrm{avail}},\,0\right\}.
\label{eq:waste}
\end{equation}
The bound follows because preparation can begin at most$T_{\max}+E_{\mathrm{early}}$ before the actual handover, whereas preparation performed during the final $T_{\mathrm{avail}}$ interval is useful. 
Equation~\eqref{eq:waste} bounds stale-preparation duration, rather than GPU cycles, transferred bytes, or occupied VRAM.

\subsection{Contention-Aware Migration Operation}
\label{sec:contention}

\method does not assume exclusive access to the target GPU or inter-gNB link. 
Instead, the Scheduler Engine uses the effective prefill throughput $r_p$ and goodput $B$ available to each migration session, allowing the selected window to reflect current resource contention.
Under handover--handover contention, concurrent prefix-prefill tasks share the target GPU and their suffix streams share the inter-gNB link. 
The resulting effective rates are incorporated into the periodic window decision, while target-side admission prevents the aggregate inference and migration state from exceeding available VRAM. 
If contention increases after preparation begins, migration continues at the reduced rates and any remaining work contributes to SIT.
Under handover--inference contention, migration prefill competes with resident decoding requests. 
\method uses chunked prefill to interleave prefix reconstruction with decoding and limit long GPU-blocking operations \cite{agrawal2023sarathi}. 
Incremental KV-cache extraction and transfer proceed asynchronously, while their memory and interconnect overhead remains reflected in the measured effective rates.
\method adapts when migration begins but does not jointly allocate GPU cycles or link bandwidth among competing requests. 
Such allocation remains the responsibility of the underlying serving and network runtimes.

\section{System Implementation} \label{sec:implementation}
We implement \method on vLLM 0.8.5~\cite{kwon2023efficient} with approximately 1.7K lines of Python code.
Each gNB runs an independent vLLM \texttt{AsyncLLMEngine}, while a node daemon coordinates migration through asynchronous control requests and gRPC-based KV-cache transfer.
Rather than modifying model kernels, \method directly interfaces with vLLM's scheduler, PagedAttention block manager, and GPU KV cache to access and reconstruct request state.
We use a PagedAttention block size of 16 tokens and the same implementation for the single-host and cross-host experiments in \S~\ref{sec:exp_set}.

\noindent\textbf{Parallel state preparation.}
We align the prefix--suffix split to PagedAttention block boundaries so that only completed KV blocks are transferred during decoding.
At the trigger, the source identifies the last complete block, snapshots the prefix token IDs, and sends them to the target for normal vLLM prefill.
The source continues decoding without waiting for the target.
As complete blocks are produced, \method obtains their physical block IDs from vLLM's block manager, extracts the corresponding per-layer KV tensors from the GPU cache, and asynchronously streams them through gRPC.
The tensors are serialized in memory using PyTorch and \texttt{io.BytesIO}, while the target buffers received suffix blocks for each request and independently reconstructs the prefix.

\noindent\textbf{Final synchronization and activation.}
At handover, the source takes a final token-sequence snapshot and transfers any remaining unsynchronized KV blocks before stopping source-side decoding.
Migrated KV blocks cannot be directly attached to an arbitrary active vLLM request because its scheduler state, sequence metadata, block table, and physical KV allocation must remain consistent.
\method therefore reconstructs a target-side request for the final token sequence, allocates a new physical block table, and copies the reconstructed prefix and buffered suffix blocks into the corresponding GPU locations.
It then updates the computed-token state and decoding stage and inserts the request into vLLM's running queue, allowing decoding to resume from position $n$ without repeating prefill over $[0,n)$.

\noindent\textbf{Cancellation and contention.}
\method maintains migration state separately for each request, including the selected target, transfer progress, and buffered blocks, so cancelled preparation can be discarded without affecting other requests.
KV extraction and transfer execute asynchronously with source-side generation, while target-side prefix reconstruction and resource sharing remain managed by the underlying vLLM and communication runtimes.

\section{Performance Evaluation}\label{sec:evaluation}
\subsection{Experimental Methodology}
\label{sec:exp_set}

\begin{figure*}[t]
    \centering
    \includegraphics[width=\textwidth]{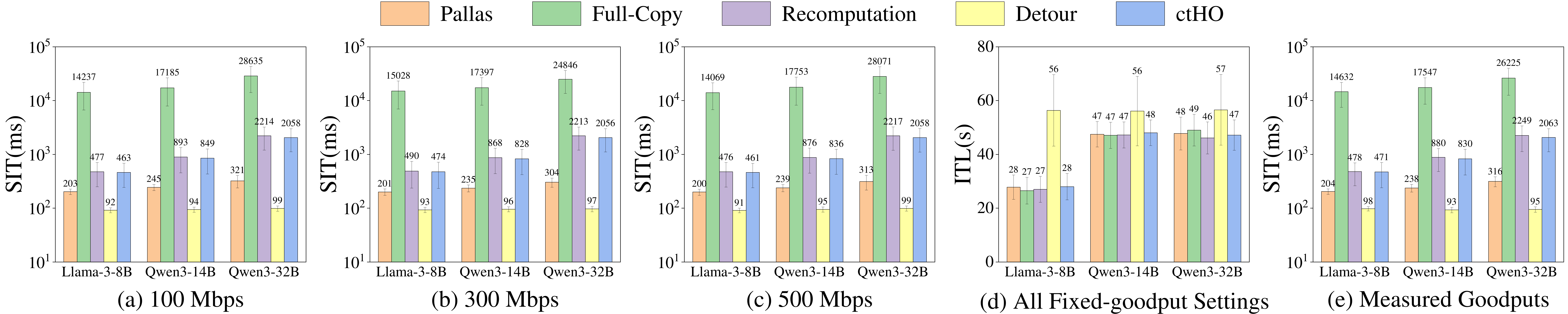}
    \caption{Single-handover performance of five methods across three models:
    (a)--(c) SIT under fixed goodputs of 100, 300, and 500~Mbps;
    (d) ITL across all fixed-goodput settings;
    (e) SIT under measured goodput traces.}
    \label{fig:exp1}
    \vspace{-5mm}
\end{figure*}

\textbf{Hardware and Workloads.}
For the single-host experiments, the source and target gNBs are deployed on a server with an Intel(R) Xeon(R) Platinum 8358P CPU (2.60~GHz), 503~GiB of memory, and NVIDIA RTX A6000 GPUs (48~GB each), with each gNB exclusively using one GPU.
For the cross-host experiment, the two gNBs run on two physical servers connected by 1-Gbps full-duplex Ethernet, each using one RTX A6000 GPU.
The native inter-host RTT is 0.21~ms and the unshaped TCP goodput is approximately 940~Mbps.
All experiments use our vLLM 0.8.5 prototype.

We evaluate Meta-Llama-3-8B, Qwen3-14B, and Qwen3-32B-Instruct-AWQ using ShareGPT conversational requests.
The AWQ-quantized checkpoint allows Qwen3-32B to fit within a single RTX A6000.
For the single-handover and ablation experiments, we divide the 500--4,500-token context range into four equal intervals and randomly select 15 samples from each interval.
Repeated-handover sessions start with a 1,000-token prompt and generate tokens throughout trajectory replay.
The concurrency experiment uses a fixed 2,000-token context, while the window-decision and prediction-stress experiments use 1,000-, 2,000-, or 3,000-token contexts.

\textbf{Network and Mobility Settings.}
The controlled experiments use inter-gNB goodputs of 100, 300, and 500~Mbps, covering bandwidth-constrained operating regimes~\cite{ngmn2012smallcellbackhaul}. 
Within each setting, all compared schemes use the same goodput.
We additionally replay time-varying goodput traces from the SNOB-5G dataset~\cite{ferreira2024experimental}, collected from an outdoor 60-GHz WiGig testbed under normal and blockage conditions, over the native cross-host path.
We use vehicle trajectories from the nuScenes v1.0-trainval split~\cite{Caesar_2020_CVPR} and integrate a constant-velocity-and-heading (CVH) predictor~\cite{karle2024selfevaluation}. 
Using the UE's latest position, velocity, and heading, it predicts the future trajectory, target gNB, and handover time. 
We evaluate its prediction availability and impact on \method in \S~\ref{Efficiency}.

Trajectories are translated and rotated, without scaling, into square-grid topologies with cell side lengths of 50, 100, and 150~m. 
We linearly interpolate between consecutive samples that cross a cell boundary to obtain the actual handover time and derive $T_{\mathrm{avail}}$ as defined in \S~\ref{sec:runtime_replanning}.
For the single-handover, window-decision, ablation, and prediction-stress experiments, we extract individual handover events and drive the prototype using their predicted targets and handover times. 
Each paired comparison uses the same event, predictor outputs, workload, and network setting. 
The prediction-stress experiment varies $T_{\mathrm{avail}}$ by delaying the stable correct target prediction. 
The repeated-handover experiment instead replays complete trajectory segments with the predictor and Scheduler Engine operating online.

\textbf{Baselines.}
We compare \method with \Detour, \FC, \RC, and \ctHO under identical workloads, network conditions, and handover events.
For \ctHO, the prefix--suffix split is selected at KV-block granularity to minimize post-handover recovery delay.
We use its single-user formulation for the single- and
repeated-handover experiments and its multi-user formulation
when multiple UEs hand over concurrently to the same target,
under the same aggregate inter-gNB goodput constraint.

\textbf{Metrics.}
We report service interruption time (SIT) and inter-token latency (ITL) as defined in \S~\ref{sec:bottlenecks}.
The window-decision experiment additionally reports the early-preparation exposure $T_{\mathrm{early}}$ defined in Eq.~\eqref{eq:early}.
We use \(T_{\mathrm{avail}}\) to characterize the prediction time available before handover. 

\begin{table}[t]
\centering
\caption{Repeated-handover SIT and ITL performance across three models.}
\label{tab:exp4}
\scriptsize
\setlength{\tabcolsep}{2.2pt}
\renewcommand{\arraystretch}{0.92}
\begin{tabular}{llrrrrr}
\toprule
\textbf{Model} & \textbf{Metric} & \textbf{\Detour} & \textbf{\FC} & \textbf{\RC} & \textbf{\ctHO} & \textbf{\method} \\
\midrule

\multirow{7}{*}{Llama-3-8B}
& SIT Avg. & 287.7 & 19645.7 & 837.3 & 812.7 & 153.8 \\
& SIT P50  & 279.1 & 18018.7 & 828.8 & 812.3 & 152.8 \\
& SIT P90  & 481.7 & 26931.4 & 1082.8 & 1040.2 & 190.9 \\
& SIT P99  & 652.2 & 36163.7 & 1252.2 & 1230.3 & 216.3 \\
& ITL Avg. & 241.3 & 25.0 & 25.1 & 25.1 & 25.1 \\
& ITL P90  & 433.7 & 26.4 & 26.4 & 26.4 & 26.4 \\
& ITL P99  & 642.9 & 27.3 & 27.3 & 27.3 & 27.2 \\
\midrule

\multirow{7}{*}{Qwen3-14B}
& SIT Avg. & 307.2 & 20983.8 & 1360.8 & 1305.9 & 232.3 \\
& SIT P50  & 298.1 & 20163.3 & 1327.7 & 1273.5 & 227.7 \\
& SIT P90  & 501.4 & 26898.7 & 1676.2 & 1604.2 & 279.1 \\
& SIT P99  & 682.8 & 36087.5 & 2176.0 & 2079.5 & 343.4 \\
& ITL Avg. & 261.7 & 45.4 & 45.5 & 45.5 & 45.6 \\
& ITL P90  & 454.2 & 46.9 & 46.8 & 46.8 & 46.9 \\
& ITL P99  & 663.4 & 47.6 & 47.5 & 47.6 & 47.7 \\
\midrule

\multirow{7}{*}{Qwen3-32B}
& SIT Avg. & 311.9 & 19746.9 & 4432.8 & 3678.6 & 323.8 \\
& SIT P50  & 306.3 & 19056.7 & 4248.1 & 3621.7 & 317.9 \\
& SIT P90  & 523.2 & 24856.9 & 4891.0 & 4163.2 & 387.4 \\
& SIT P99  & 699.3 & 32802.9 & 5533.8 & 4830.8 & 488.1 \\
& ITL Avg. & 283.0 & 66.7 & 66.8 & 66.7 & 66.7 \\
& ITL P90  & 475.4 & 68.0 & 68.0 & 68.0 & 68.0 \\
& ITL P99  & 684.6 & 68.8 & 69.1 & 68.9 & 68.8 \\
\bottomrule

\end{tabular}
\end{table}

\subsection{Effectiveness of \method}
\label{sec:Effectiveness}

\textbf{Single-Handover Performance.}
We first compare the five schemes under fixed inter-gNB goodputs. 
As shown in Figures~\ref{fig:exp1}(a)--(c), \method consistently achieves the lowest average SIT among the target-side recovery schemes. 
Even at 100~Mbps with Qwen3-32B, its average SIT is only 321~ms. 
Across all settings, \method reduces average SIT by factors of 70.13--89.68, 2.35--7.28, and 2.28--6.76 relative to \FC, \RC, and \ctHO, respectively. 
These gains arise because the baselines begin state recovery after handover, whereas \method completes most prefix reconstruction and suffix synchronization beforehand.
Figure~\ref{fig:exp1}(d) shows that \method maintains an average ITL of 28--48~ms, reducing it by 16.0\%--50.0\% relative to \Detour by restoring serving locality after handover.
Under the measured time-varying goodput traces in Figure~\ref{fig:exp1}(e), \method achieves an average SIT of 204--316~ms and reduces SIT by factors of 71.73--82.99, 2.34--7.12, and 2.31--6.53 relative to \FC, \RC, and \ctHO, respectively. 
These results confirm that its benefits persist across physical hosts and time-varying network conditions.

\textbf{Repeated-Handover Performance.}
We replay nuScenes trajectory segments over an $8\times8$ grid of 100-m cells, with the CVH predictor and Scheduler Engine operating throughout each session. 
We conduct ten rounds, each containing eight trajectories for every handover count from one to eight. 
This produces 64 sessions and 288 handovers per round, or 2,880 handovers per model. 
For each handover, goodput is sampled from a normal distribution with a mean of 300~Mbps and a standard deviation of 50~Mbps, truncated below at 150~Mbps. 
All schemes use identical trajectories, predictions, goodput samples, and workloads.
As shown in Table~\ref{tab:exp4}, \method achieves an average SIT of 153.8--323.8~ms and a P99 SIT of 216.3--488.1~ms across the three models, outperforming all target-side recovery baselines. 
Its average and P99 ITLs remain at 25.1--66.7~ms and 27.2--68.8~ms, respectively, while \Detour incurs 4.24--9.60$\times$ its average ITL because repeated handovers progressively lengthen the forwarding path. 
These results demonstrate that \method preserves low SIT and stable token delivery across long-running sessions with repeated mobility events.

\subsection{Window Decision and Prediction Robustness}
\label{Efficiency}

\begin{figure}[t]
    \centering
    \includegraphics[width=\linewidth]{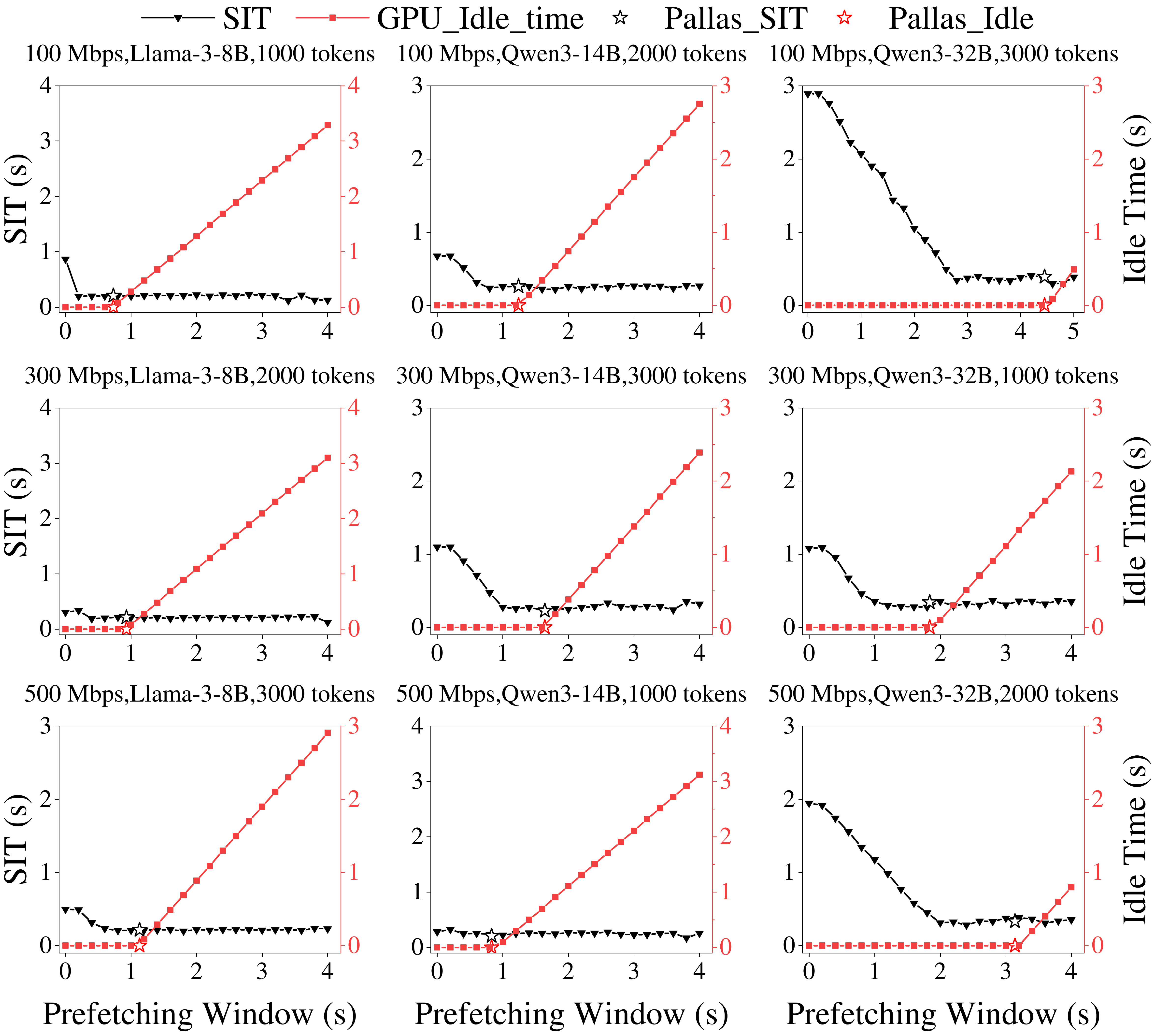}
    \caption{Prefetching-window selection across different goodputs, models, and context lengths.}
    \label{fig:exp2}
    \vspace{-5mm}
\end{figure}

\textbf{Prefetching-Window Decision.}
We evaluate the Scheduler Engine by sweeping fixed windows from 0 to 5~s in 0.2-s increments and comparing them with the window selected by \method. 
We use nine configurations from a three-level orthogonal array spanning inter-gNB goodput, model size, and context length. 
Within each configuration, the handover event, predictor outputs, and other settings remain unchanged.
Figure~\ref{fig:exp2} confirms the trade-off modeled in \S~\ref{Sec:modeling}: short windows leave more preparation unfinished at handover, whereas long windows increase early-preparation exposure without further reducing SIT. 
For example, at 300~Mbps with Qwen3-14B and a 3,000-token context, \method selects a 1.64-s window, resulting in a SIT of 227.57~ms and only 0.017~s of early-preparation exposure. 
Across all nine configurations, the selected windows keep both metrics near their low-value regions. We set $\alpha=0.8$ in Eq.~\eqref{eq:cost}, prioritizing SIT reduction over early preparation.

\begin{figure}[t]
    \centering
    \includegraphics[width=\linewidth]{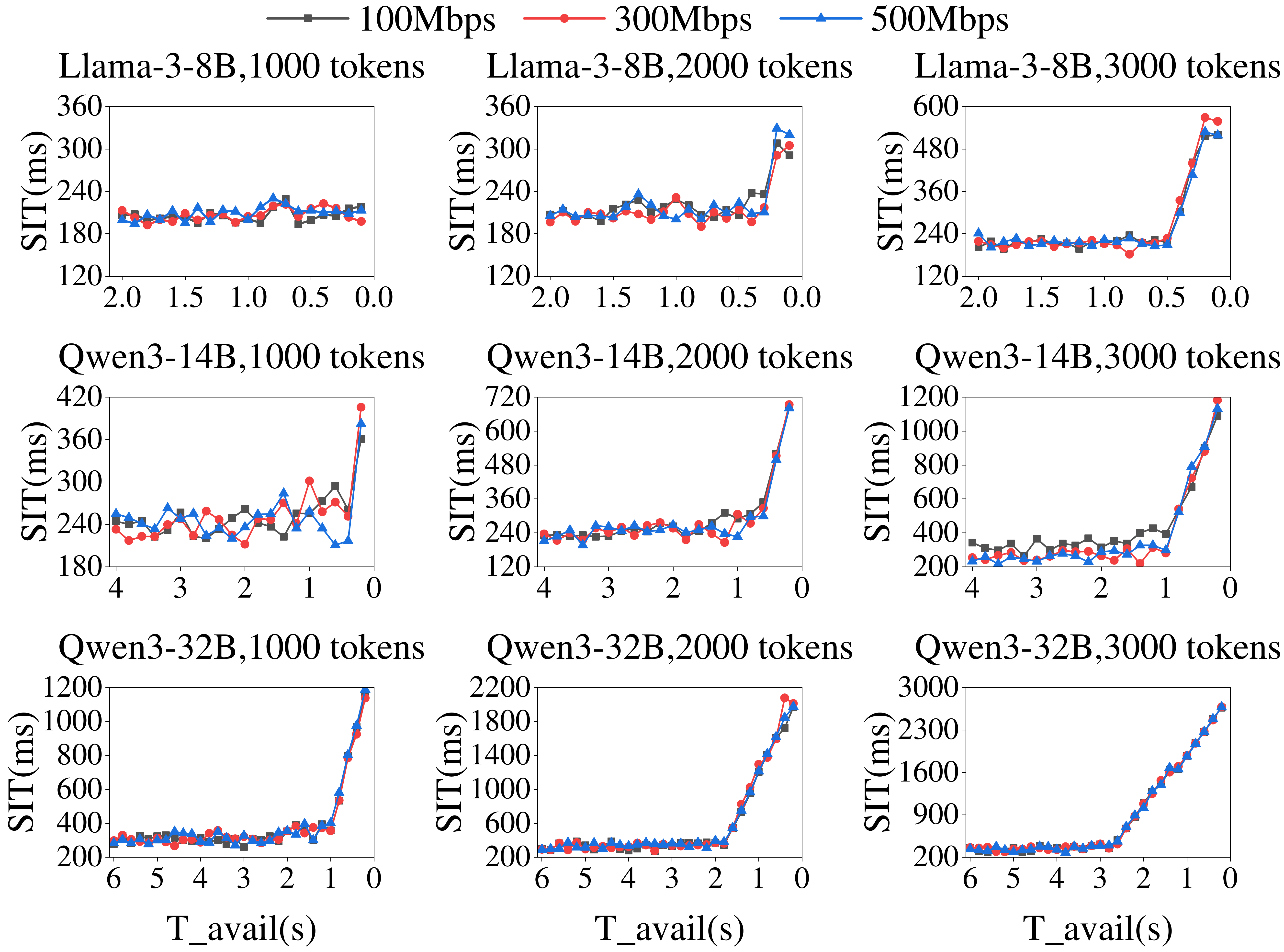}
    \caption{Performance of \method under controlled reductions in $T_{\mathrm{avail}}$.}
    \label{fig:exp3}
    \vspace{-5mm}
\end{figure}

\noindent\textbf{Prediction robustness.}
Table~\ref{tab:prediction} characterizes the CVH predictor used in our evaluation. 
As the cell size increases from 50 to 150~m, the fraction of handovers with at least 2~s of $T_{\mathrm{avail}}$ increases from 51.9\% to 62.8\%, while the fraction experiencing an incorrect target prediction decreases from 37.3\% to 11.2\%. 
The P90 $E_{\mathrm{early}}$ also decreases from 4.19 to 0.94~s. Correspondingly, the mean $T_{\mathrm{waste}}$ decreases from 1.46 to 0.58~s and remains below the mean analytical bound for every cell size.
We further stress \method by varying $T_{\mathrm{avail}}$ from 0 to 6~s in 0.1-s increments while keeping each handover event, workload, and network setting unchanged. 
As shown in Figure~\ref{fig:exp3}, SIT remains low until the available preparation time becomes insufficient for the selected model and context. For example, at 300~Mbps with Qwen3-14B and a 2,000-token context, noticeable degradation appears only below $T_{\mathrm{avail}}=1$~s.
Reducing $T_{\mathrm{avail}}$ leaves more preparation on the post-handover path and therefore increases SIT. 
Together, Figure~\ref{fig:exp3} and Table~\ref{tab:prediction} show that prediction availability determines the achievable benefit, while stale-preparation duration remains consistent with the bound in \S~\ref{sec:runtime_replanning}.

\begin{table}[t]
\centering
\caption{Prediction availability and failure overhead.}
\label{tab:prediction}
\scriptsize
\setlength{\tabcolsep}{3.0pt}
\renewcommand{\arraystretch}{1.05}
\begin{tabular}{@{}c c c c c c c@{}}
\toprule
\shortstack{Cell\\(m)}
&
\shortstack{$T_{\mathrm{avail}}\!\geq\!2$ s\\(\%)}
&
\shortstack{$T_{\mathrm{avail}}\!\geq\!3$ s\\(\%)}
&
\shortstack{$\hat g_{\mathrm{tar}}\neq g^*_{\mathrm{tar}}$\\(\%)}
&
\shortstack{$E_{\mathrm{early}}$\\P90 (s)}
&
\shortstack{$T_{\mathrm{waste}}$\\Mean (s)}
&
\shortstack{Upper Bound\\Mean (s)}
\\
\midrule
50  & 51.9 & 34.6 & 37.3 & 4.19 & 1.46 & 3.79 \\
100 & 57.9 & 42.2 & 19.3 & 2.16 & 1.02 & 2.93 \\
150 & 62.8 & 48.0 & 11.2 & 0.94 & 0.58 & 2.57 \\
\bottomrule
\end{tabular}
\end{table}

\subsection{Concurrency Evaluation}
\label{sec:Concurrency}

\begin{figure}[t]
    \centering
    \includegraphics[width=\linewidth]{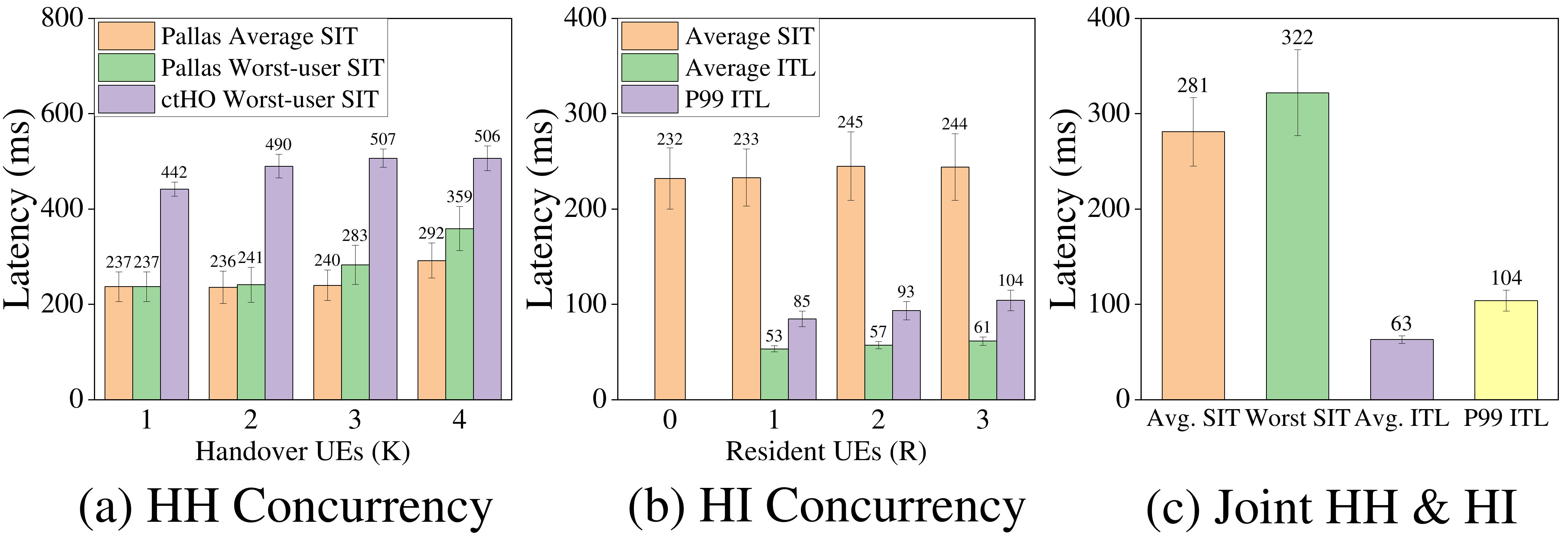}
    \caption{Concurrency performance under (a) HH, (b) HI, and (c) joint HH--HI contention.}
    \label{fig:concurrency}
    \vspace{-5mm}
\end{figure}

We evaluate \method under handover--handover (HH) and handover--inference (HI) contention, as well as a joint HH--HI case. HH represents concurrent migrations to the same target, whereas HI represents migration preparation sharing the target GPU with resident inference. All experiments use Qwen3-14B, a 2,000-token migration context, and an aggregate inter-gNB goodput of 1~Gbps. Each configuration is repeated ten times; error bars in Figure~\ref{fig:concurrency} show the corresponding standard deviations.

\noindent\textbf{Handover--handover contention.}
We simultaneously migrate $K\in{1,2,3,4}$ UEs from one source to the same target, with $K=1$ serving as the non-concurrent reference. 
Their prefill tasks share the target GPU in first-come, first-served order, while their KV-cache transfers equally share the aggregate goodput. 
We compare \method with multi-user \ctHO under the same resource constraints and report average and worst-user SIT.
As shown in Figure~\ref{fig:concurrency}(a), the average SIT of \method remains at 236--240~ms for $K\leq3$ and increases to 292~ms at $K=4$. Its worst-user SIT increases from 237~ms at $K=1$ to 283~ms at $K=3$ and 359~ms at $K=4$. 
In comparison, the worst-user SIT of \ctHO is 442--507~ms, with \method reducing it by 29--51\%. Proactive preparation absorbs most of the contention before handover; only at higher concurrency does the increased prefill and transfer backlog noticeably raise SIT. In contrast, \ctHO exposes its concurrent recovery operations entirely after handover.

\noindent\textbf{Handover--inference contention.}
We next migrate one UE to a target serving $R\in{0,1,2,3}$ resident sessions, where $R=0$ is the no-load reference. Migration prefill and resident decoding share the GPU, and a 256-token prefill chunk is used to interleave their execution. Because only one UE migrates, its transfer uses the full configured inter-gNB goodput.
Figure~\ref{fig:concurrency}(b) shows that the migrating UE's average SIT increases only from 232~ms at $R=0$ to 244~ms at $R=3$. Across $R=1$--3, the resident UEs' average ITL increases from 53.24 to 61.38~ms, while their P99 ITL rises from 84.67 to 104.15~ms. Chunked prefill therefore limits the effect on average decoding latency, although the P99 result at $R=3$ reveals measurable tail interference under heavier target-side load.

\noindent\textbf{Joint HH--HI contention.}
Finally, we evaluate $K=2$ migrating UEs sharing the target with $R=2$ resident sessions. As shown in Figure~\ref{fig:concurrency}(c), \method achieves an average SIT of 281~ms and a worst-user SIT of 322~ms. The resident UEs maintain an average ITL of 63~ms, while their P99 ITL reaches 104~ms. The simultaneous GPU and link contention increases both migration and inference latency relative to the isolated cases, but average migration SIT remains below 300~ms.

Overall, \method maintains an average SIT below 300~ms across all evaluated HH, HI, and joint configurations, while contention primarily affects worst-user SIT and resident tail ITL. Joint multi-user resource allocation remains outside the scope of this work.

\subsection{Ablation Study}
\label{sec:Ablation_study}

We isolate the benefits of historical KV-cache recomputation and incremental KV-cache streaming using two variants of \method.
To control for proactive timing, all variants use the same handover events, predictor outputs, prefetching-window decisions, workloads, and network conditions.
\method w/o RC disables historical-prefix reconstruction and proactively transfers the complete KV cache from the source.
\method w/o SM disables incremental streaming; after handover, the target receives the tokens generated during the prefetching window and recomputes their KV cache before resuming decoding.

\begin{figure}[t]
    \centering
    \includegraphics[width=\linewidth]{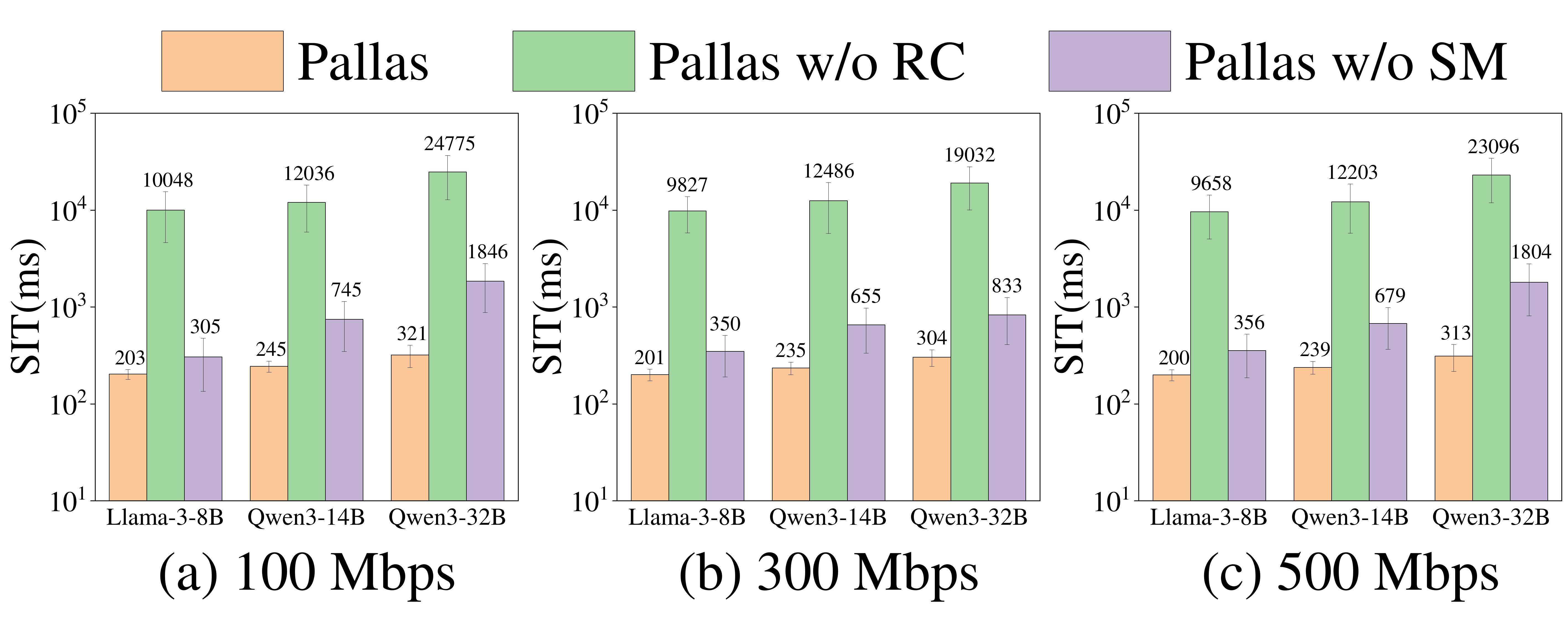}
    \caption{Ablation study of single-handover SIT across three goodputs and three models.}
    \label{fig:Ablation_study}
    \vspace{-5mm}
\end{figure}

As shown in Figure~\ref{fig:Ablation_study}, removing either component increases SIT.
At 300~Mbps with Qwen3-14B, \method achieves an SIT of 235~ms, compared with 12,486~ms for \method w/o RC and 655~ms for \method w/o SM.
Across all evaluated settings, \method reduces SIT by an average of 97.93\% and 33.3\% relative to the two variants, respectively.
Without recomputation, transferring the complete KV cache remains communication-intensive; without incremental streaming, the target must recompute the state generated during preparation after handover.
Combining the two mechanisms therefore minimizes the residual work on the post-handover interruption path.

\section{Related Work}\label{sec:related}

Service migration has been extensively studied in mobile edge computing (MEC). Prior systems use Markov decision processes, deep reinforcement learning, mobility prediction, and collaborative optimization to determine when and where services should migrate \cite{taleb2016follow,mwasinga2023rasm,zhao2025mapsm,zeng2025towards}. These approaches generally model service state as a relatively static transferable object, whereas an active LLM request maintains a KV cache that grows throughout decoding. Data-center LLM serving systems address this state through direct KV-cache migration, as in Llumnix \cite{sun2024llumnix}, or token-based recomputation, as in ServerlessLLM \cite{fu2024serverlessllm}. KV-cache compression techniques, including CacheGen, GEAR, and KIVI, can further reduce storage or transfer volume \cite{10.1145/3651890.3672274,Hao2024Gear,liu2024kivi}. However, these systems do not consider mobility-induced handover or the need to prepare an evolving inference state while source-side decoding continues.

AI-RAN brings inference into radio access networks, motivating mobility support for edge AI workloads. AoRA demonstrates AI inference colocated with RAN functions \cite{kholmatov2025aora}, while the AI-RAN working group highlights proactive mobility management as an important requirement \cite{ai_ran_wg3_2026}. The most closely related work, \ctHO \cite{lee2026lowlatencyedgellmhandover}, combines target-side token prefill with direct KV-cache transfer and jointly optimizes their partition and backhaul allocation to reduce worst-user handover delay. However, both recovery operations begin after handover. In contrast, \method prepares the target beforehand by overlapping historical-prefix reconstruction and incremental KV-cache streaming with ongoing source-side inference, and adaptively selects the prefetching window to reduce the work remaining on the post-handover interruption path.

\section{Conclusion}\label{sec:conclusion} 
In this work, we present \method, a proactive parallel KV-cache migration framework for mobile LLM serving in AI-RAN.By overlapping target-side recomputation and incremental KV-cache streaming with ongoing source-side inference, \method moves most state-preparation work before handover.Our vLLM-based prototype and trace-driven evaluation show that \method substantially reduces SIT compared with target-side recovery baselines while avoiding the persistent ITL penalty of \Detour.Future work will explore joint resource allocation and window replanning under dynamic multi-user contention.

\bibliographystyle{unsrt}
\bibliography{main}

@inproceedings{abdah2020handover,
  title={Handover Prediction Integrated with Service Migration in {5G} Systems},
  author={Abdah, Hadeel and Barraca, Jo{\~a}o Paulo and Aguiar, Rui L.},
  booktitle={ICC 2020 - 2020 IEEE International Conference on Communications (ICC)},
  pages={1--7},
  year={2020},
  doi={10.1109/ICC40277.2020.9149426}
}

@article{mei2022realtime,
  title={Realtime Mobile Bandwidth and Handoff Predictions in {4G/5G} Networks},
  author={Mei, Lifan and Gou, Jinrui and Cai, Yujin and Cao, Houwei and Liu, Yong},
  journal={Computer Networks},
  volume={204},
  pages={108736},
  year={2022},
  doi={10.1016/j.comnet.2021.108736}
}

@inproceedings{ahmad2025edgewarp,
  title={{Warping the Edge}: Enabling Instant Mobility for Stateful Applications over {5G} and Beyond},
  author={Ahmad, Mukhtiar and Bilal, Faaiq and Ali, Mutahar and Nawazish, Syed Muhammad Ali and Salman, Amir and Ali, Shazer and Ahmad, Fawad and Qazi, Zafar Ayyub},
  booktitle={Proceedings of the Tenth ACM/IEEE Symposium on Edge Computing},
  pages={1--18},
  year={2025},
  publisher={Association for Computing Machinery},
  doi={10.1145/3769102.3770610}
}

@article{kundu2025ai,
  title={{AI-RAN}: Transforming {RAN} with {AI}-Driven Computing Infrastructure},
  author={Kundu, Lopamudra and Lin, Xingqin and Gadiyar, Rajesh and Lacasse, Jean-Francois and Chowdhury, Shuvo},
  journal={arXiv preprint arXiv:2501.09007},
  year={2025}
}

@inproceedings{kholmatov2025aora,
  title={{AoRA}: {AI-on-RAN} for Backhaul-Free Edge Inference},
  author={Kholmatov, Siyavushkhon and Cho, Seongsik and Chong, Song and Lee, Kyunghan},
  booktitle={Proceedings of the ACM SIGCOMM 2025 Conference},
  pages={1263--1265},
  year={2025}
}

@article{taleb2016follow,
  title={Follow-me cloud: When cloud services follow mobile users},
  author={Taleb, Tarik and Ksentini, Adlen and Frangoudis, Pantelis A},
  journal={IEEE Transactions on Cloud Computing},
  volume={7},
  number={2},
  pages={369--382},
  year={2016}
}

@article{mwasinga2023rasm,
  title={{RASM}: Resource-Aware Service Migration in Edge Computing Based on Deep Reinforcement Learning},
  author={Mwasinga, Lusungu Josh and Le, Duc-Tai and Raza, Syed M and Challa, Rajesh and Kim, Moonseong and Choo, Hyunseung},
  journal={Journal of Parallel and Distributed Computing},
  volume={182},
  pages={104745},
  year={2023}
}

@article{10024837,
  author={Polese, Michele and Bonati, Leonardo and D'Oro, Salvatore and Basagni, Stefano and Melodia, Tommaso},
  journal={IEEE Communications Surveys \& Tutorials}, 
  title={Understanding {O-RAN}: Architecture, Interfaces, Algorithms, Security, and Research Challenges}, 
  year={2023},
  volume={25},
  number={2},
  pages={1376--1411},
  doi={10.1109/COMST.2023.3239220}
}

@inproceedings{fu2024serverlessllm,
  title={{ServerlessLLM}: {Low-Latency} Serverless Inference for Large Language Models},
  author={Fu, Yao and Xue, Leyang and Huang, Yeqi and Brabete, Andrei-Octavian and Ustiugov, Dmitrii and Patel, Yuvraj and Mai, Luo},
  booktitle={18th USENIX Symposium on Operating Systems Design and Implementation (OSDI 24)},
  pages={135--153},
  year={2024}
}

@inproceedings{10.1145/3651890.3672274,
author = {Liu, Yuhan and Li, Hanchen and Cheng, Yihua and Ray, Siddhant and Huang, Yuyang and Zhang, Qizheng and Du, Kuntai and Yao, Jiayi and Lu, Shan and Ananthanarayanan, Ganesh and Maire, Michael and Hoffmann, Henry and Holtzman, Ari and Jiang, Junchen},
title = {{CacheGen}: {KV} Cache Compression and Streaming for Fast Large Language Model Serving},
year = {2024},
isbn = {9798400706141},
publisher = {Association for Computing Machinery},
address = {New York, NY, USA},
doi = {10.1145/3651890.3672274},
booktitle = {Proceedings of the ACM SIGCOMM 2024 Conference},
pages = {38--56},
location = {Sydney, NSW, Australia},
series = {ACM SIGCOMM '24}
}

@techreport{ai_ran_wg3_2026,
  author      = {{AI-RAN Alliance}},
  title       = {{AI-on-RAN}: Enabling Monetizable Differentiated Connectivity for {AI}},
  institution = {AI-RAN Alliance},
  year        = {2026},
  type        = {Working Group 3 White Paper},
  url         = {https://ai-ran.org/documents/AI-RAN-WG3-AI-on-RAN-Whitepaper.pdf},
  lastaccessed = {April 1, 2026}
}

@article{agrawal2023sarathi,
  title={{Sarathi}: Efficient {LLM} Inference by Piggybacking Decodes with Chunked Prefills},
  author={Agrawal, Amey and Panwar, Ashish and Mohan, Jayashree and Kwatra, Nipun and Gulavani, Bhargav S and Ramjee, Ramachandran},
  journal={arXiv preprint arXiv:2308.16369},
  year={2023}
}

@article{zhao2025mapsm,
  title={{MAPSM}: Mobility-Aware Proactive Service Migration Framework for Mobile Edge Computing in Consumer Internet of Vehicles},
  author={Zhao, Xuhui and Shi, Yan and Chen, Shanzhi and Liu, Jianghui and Ji, Baofeng and Mumtaz, Shahid},
  journal={IEEE Transactions on Consumer Electronics},
  year={2025}
}

@article{zeng2025towards,
  title={Towards Collaborative and Latency-Aware Microservice Migration in Mobile Edge Computing},
  author={Zeng, Luchuan and Zhang, Chen and Wang, Zichen and Du, Hongwei and Jia, Xiaohua},
  journal={IEEE Internet of Things Journal},
  year={2025}
}

@article{yang2025qwen3,
  title={{Qwen3} Technical Report},
  author={Yang, An and Li, Anfeng and Yang, Baosong and Zhang, Beichen and Hui, Binyuan and Zheng, Bo and Yu, Bowen and Gao, Chang and Huang, Chengen and Lv, Chenxu and others},
  journal={arXiv preprint arXiv:2505.09388},
  year={2025}
}

@inproceedings{kwon2023efficient,
  title={Efficient Memory Management for Large Language Model Serving with {PagedAttention}},
  author={Kwon, Woosuk and Li, Zhuohan and Zhuang, Siyuan and Sheng, Ying and Zheng, Lianmin and Yu, Cody Hao and Gonzalez, Joseph and Zhang, Hao and Stoica, Ion},
  booktitle={Proceedings of the 29th symposium on operating systems principles},
  pages={611--626},
  year={2023}
}

@inproceedings{sun2024llumnix,
  title={{Llumnix}: Dynamic Scheduling for Large Language Model Serving},
  author={Sun, Biao and Huang, Ziming and Zhao, Hanyu and Xiao, Wencong and Zhang, Xinyi and Li, Yong and Lin, Wei},
  booktitle={18th USENIX symposium on operating systems design and implementation (OSDI 24)},
  pages={173--191},
  year={2024}
}

@inproceedings{Hao2024Gear,
  title = {{GEAR}: An Efficient Error Reduction Framework for {KV} Cache Compression in {LLM} Inference},
  author = {Kang, Hao and Zhang, Qingru and Kundu, Souvik and Jeong, Geonhwa and Liu, Zaoxing and Krishna, Tushar and Zhao, Tuo},
  booktitle = {Proceedings of The 4th NeurIPS Efficient Natural Language and Speech Processing Workshop},
  pages = {305--321},
  year = {2024},
  volume = {262},
}

@article{liu2024kivi,
  title={{KIVI}: A Tuning-Free Asymmetric 2-Bit Quantization for {KV} Cache},
  author={Liu, Zirui and Yuan, Jiayi and Jin, Hongye and Zhong, Shaochen and Xu, Zhaozhuo and Braverman, Vladimir and Chen, Beidi and Hu, Xia},
  journal={arXiv preprint arXiv:2402.02750},
  year={2024}
}

@inproceedings{298687,
  author = {Yinmin Zhong and Shengyu Liu and Junda Chen and Jianbo Hu and Yibo Zhu and Xuanzhe Liu and Xin Jin and Hao Zhang},
  title = {{DistServe}: Disaggregating Prefill and Decoding for Goodput-optimized Large Language Model Serving},
  booktitle = {18th USENIX Symposium on Operating Systems Design and Implementation (OSDI 24)},
  year = {2024},
  pages = {193--210},
}

@techreport{ngmn2012smallcellbackhaul,
  author      = {{NGMN Alliance}},
  title       = {Small Cell Backhaul Requirements},
  institution = {Next Generation Mobile Networks Alliance},
  type        = {White Paper},
  number      = {Version 1.0 Final},
  year        = {2012},
  month       = jun,
  url         = {https://ngmn.org/wp-content/uploads/NGMN_Whitepaper_Small_Cell_Backhaul_Requirements.pdf},
  note        = {Final deliverable, approved 4 June 2012}
}

@misc{lee2026lowlatencyedgellmhandover,
      title={Low-Latency Edge {LLM} Handover via Joint {KV} Cache Transfer and Token Prefill}, 
      author={Seunghun Lee and Jihong Park and Ce Zheng and Hyuncheol Park},
      year={2026},
      eprint={2603.28018},
      archivePrefix={arXiv},
      primaryClass={eess.SP},
      url={https://arxiv.org/abs/2603.28018}, 
}

@inproceedings {servegen,
    author = {Yuxing Xiang and Xue Li and Kun Qian and Yan Zhang and Wenyuan Yu and Ennan Zhai and Xin Jin and Jingren Zhou},
    title = {{ServeGen}: Workload Characterization and Generation of Large Language Model Serving in Production},
    booktitle = {23rd USENIX Symposium on Networked Systems Design and Implementation (NSDI 26)},
    year = {2026},
    isbn = {978-1-939133-54-0},
    address = {Renton, WA},
    pages = {1845--1859},
    url = {https://www.usenix.org/conference/nsdi26/presentation/xiang-servegen},
    publisher = {USENIX Association},
    month = may
}

@misc{3gpp_edge_computing,
  author = {{3GPP}},
  title  = {Enabling Edge Computing Applications in 3GPP},
  year   = {2021},
  month  = feb,
  url    = {https://www.3gpp.org/news-events/3gpp-news/edge-sa6},
  note   = {Accessed: July 28, 2026}
}

@InProceedings{Caesar_2020_CVPR,
author = {Caesar, Holger and Bankiti, Varun and Lang, Alex H. and Vora, Sourabh and Liong, Venice Erin and Xu, Qiang and Krishnan, Anush and Pan, Yu and Baldan, Giancarlo and Beijbom, Oscar},
title = {nuScenes: A Multimodal Dataset for Autonomous Driving},
booktitle = {Proceedings of the IEEE/CVF Conference on Computer Vision and Pattern Recognition (CVPR)},
month = {June},
year = {2020}
}

@article{karle2024selfevaluation,
  title   = {Self-Evaluation of Trajectory Predictors for Autonomous Driving},
  author  = {Karle, Phillip and Furtner, Lukas and Lienkamp, Markus},
  journal = {Electronics},
  volume  = {13},
  number  = {5},
  pages   = {946},
  year    = {2024},
  doi     = {10.3390/electronics13050946}
}

@misc{mlperf_inference_rules,
  author       = {{MLCommons}},
  title        = {{MLPerf Inference Rules}},
  howpublished = {\url{https://github.com/mlcommons/inference_policies/blob/master/inference_rules.adoc}},
  year         = {2025},
  note         = {Llama3.1-8B conversational serving uses a 100~ms TPOT constraint}
}

@inproceedings {318391,
author = {Shan Yu and Yifan Qiao and Mingyuan Ma and Yangmin Li and Shuo Yang and Xinyuan Tong and Yang Wang and Zhiqiang Xie and Yuwei An and Shiyi Cao and Ke Bao and Deepak Vij and Xiaoning Ding and Yichen Wang and Qingda Lu and Zhong Wang and Gao Gao and Harry Xu and Junyi Shu and Jiarong Xing and Ying Sheng},
title = {Prism: {Cost-Efficient} {Multi-LLM} Serving via {GPU} Memory Ballooning},
booktitle = {20th USENIX Symposium on Operating Systems Design and Implementation (OSDI 26)},
year = {2026},
isbn = {978-1-939133-55-7},
address = {Seattle, WA},
pages = {75--93},
url = {https://www.usenix.org/conference/osdi26/presentation/yu-shan},
publisher = {USENIX Association},
month = jul
}

@article{ferreira2024experimental,
  author  = {T{\^a}nia Ferreira and Duarte Raposo and Alexandre Figueiredo and Eurico Dias and Pedro Rito and Miguel Lu{\'i}s and Susana Sargento},
  title   = {Experimental mmWave WiGig-based backhaul network dataset},
  journal = {Data in Brief},
  volume  = {52},
  pages   = {109954},
  year    = {2024},
  doi     = {10.1016/j.dib.2023.109954}
}

\appendix

\end{document}